%% file: main.tex
\pdfoutput=1
\documentclass[11pt,letterpaper]{article}

\usepackage[margin=1in]{geometry}
\usepackage[T1]{fontenc}
\usepackage{newtxtext}
\usepackage[dvipsnames]{xcolor}
\usepackage{microtype}
\usepackage[numbers,sort&compress]{natbib}
\usepackage[hyphens]{url}
\usepackage[format=plain,labelformat=simple,labelsep=period,font=small,compatibility=false]{caption}
\usepackage[font=footnotesize,skip=3pt,subrefformat=parens]{subcaption}
\usepackage[shortlabels,inline]{enumitem}
\setlist[itemize]{noitemsep,leftmargin=*,topsep=0.3em}
\setlist[enumerate]{noitemsep,leftmargin=*,topsep=0.3em}

\input{preamble}

\makeatletter
\DeclareRobustCommand\onedot{\futurelet\@let@token\@onedot}
\def\@onedot{\ifx\@let@token.\else.\null\fi\xspace}
\makeatother

\def\cf{\emph{cf}\onedot} 
 \def\vs{\emph{vs}\onedot}

\definecolor{linkblue}{rgb}{0.21,0.49,0.74}
\usepackage[breaklinks,colorlinks,allcolors=linkblue]{hyperref}
\usepackage[capitalize]{cleveref}
\crefname{section}{Sec.}{Secs.}
\Crefname{section}{Section}{Sections}
\crefname{table}{Tab.}{Tabs.}
\Crefname{table}{Table}{Tables}

\title{\textbf{PhiJEPA: Denoising Latent Predictions for\\ Visual Representation Learning}}

\author{Makoto Yamada\textsuperscript{*} \qquad Yao-Hung Hubert Tsai\\[0.3em]
Okinawa Institute of Science and Technology\\
{\small\texttt{makoto.yamada@oist.jp}}\\[0.3em]
{\small \textsuperscript{*}Corresponding author}
}
\date{}

\begin{document}
\maketitle
\input{sec/0_abstract}
\input{sec/0_teaser_noproxy}
\input{sec/1_intro_noproxy}
\input{sec/3_method_figure}
\input{sec/2_related}
\input{sec/3_method}
\input{sec/4_experiments}
\input{sec/5_worldmodel_noproxy}

\input{sec/5_conclusion}
\input{sec/6_llm_statement}

\bibliographystyle{unsrtnat}
\bibliography{main}

\input{sec/X_suppl}
\input{sec/X_vitl}

\end{document}

%% file: preamble.tex
\usepackage{cuted} %

\usepackage{currfile} %

\usepackage{caption} %

\usepackage{graphicx}
\usepackage{amsmath,amssymb}
\usepackage{booktabs}
\usepackage{enumitem} %
\usepackage{adjustbox} %
\usepackage{multirow}
\usepackage{xspace}
\input{math_commands}

\makeatletter
\renewcommand\paragraph{\@startsection{paragraph}{4}{\z@}{0.9ex \@plus 0.3ex \@minus 0.2ex}{-0.8em}{\normalfont\normalsize\bfseries}}
\makeatother
\newcommand{\method}{PhiJEPA\xspace}

\newcommand{\sg}{\operatorname{sg}}
\newcommand{\best}[1]{\textbf{#1}}

%% file: math_commands.tex
\usepackage{bm}
\providecommand{\vtheta}{{\bm{\theta}}}
\providecommand{\vxi}{{\bm{\xi}}}

\providecommand{\vu}{{\bm{u}}}
\providecommand{\vv}{{\bm{v}}}

\providecommand{\vx}{{\bm{x}}}
\providecommand{\vy}{{\bm{y}}}
\providecommand{\vz}{{\bm{z}}}

%% file: sec/0_abstract.tex
\begin{abstract}
Can a predictor trained to infer hidden image regions still help a world model plan when part of an object disappears from view and the visible context is blurred? We introduce \method, a joint-embedding predictive architecture that adapts PhiNet's two-stage prediction to masked Transformers. From sparse, photometrically corrupted context, a Transformer block predicts hidden tokens of an independently corrupted teacher view. A token-wise MLP then maps these predictions to clean teacher tokens. Controlled ablations show that the combined objective improves dense features over the tested one-stage alternatives.
Pre-trained on ImageNet-1K with $5.6\times10^{19}$ training FLOPs, \method achieves 69.3 mIoU on linear PASCAL VOC segmentation, exceeding the evaluated I-JEPA and DSeq-JEPA checkpoints in segmentation accuracy, with 25\% fewer training FLOPs than I-JEPA. It also improves hidden-region segmentation over Simple I-JEPA (49.1 vs.\ 44.3 mIoU).
As an application, we use the frozen encoder in DINO-WM on PushT, and reuse the pre-training predictor, which JEPAs usually discard, to complete occluded tokens without fine-tuning. In open-loop planning with simultaneous object occlusion and Gaussian blur ($\sigma=2$ pixels), predictor completion raises success from 33\% to 65\%, compared with 25\% for Simple I-JEPA with its predictor and 40\% for iBOT with pixel inpainting. Together, these results support two-stage latent prediction as an efficient approach to learning visual representations that remain useful under photometric degradation and support completion under partial observation.
\end{abstract}

%% file: sec/0_teaser_noproxy.tex
\begin{figure}[t]
\centering
\includegraphics[width=\linewidth]{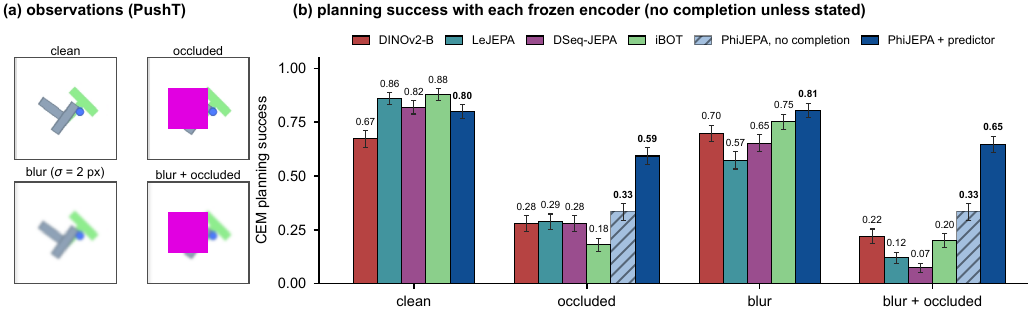}
\caption{\textbf{Planning with partially observed and degraded inputs.}
  (a)~PushT start observations: clean, occluded (magenta), blurred ($\sigma=2$ pixels) and both; the goal image is clean.
  (b)~Open-loop CEM planning success of DINO-WM models trained with the same training schedule and batch size on frozen encoders. Other encoders are shown without completion (LeJEPA has no hidden-token predictor, the public DSeq-JEPA checkpoint contains only the encoder and iBOT has no separate predictor); \method + predictor completes occluded tokens with its own predictor, which is not used without an occluder. Error bars: binomial s.e.}
\label{fig:teaser}
\end{figure}

%% file: sec/1_intro_noproxy.tex
\section{Introduction}
\label{sec:intro}

World models help agents understand their surroundings, anticipate outcomes and choose actions.
Progress in learned control and visual planning has made them an important component of physical AI~\cite{hafner2025mastering,zhou2025dino,assran2025v}.
Their effectiveness depends on the visual representation used for prediction and planning.
DINO-WM~\cite{zhou2025dino}, for example, operates in the patch-token space of a frozen pre-trained encoder.
Yet real observations are often incomplete: a robot's arm can cover an object, people can pass in front of the camera, and sensors can fail.
Moreover, observations can be degraded, for example by motion blur, low light or sensor noise.
If these changes distort the encoded state, they can make the planning cost unreliable even when the task-relevant state is unchanged.
In this paper, we ask how to pre-train, at a practical computational cost, a visual encoder whose representations are accurate and remain reliable under partial and degraded observation, and whether such an encoder benefits latent planning.

One promising approach is to use joint-embedding predictive architectures (JEPAs), which learn by predicting representations rather than reconstructing every detail of the input~\cite{path}.
For example, I-JEPA predicts teacher features of hidden image regions from visible context~\cite{assran2023self}; V-JEPA and V-JEPA~2 extend latent prediction to video and action-conditioned planning~\cite{bardes2024revisiting,assran2025v}.
These methods demonstrate the value of predicting target features from context, and sparse student inputs can reduce computation.
However, the complete training recipe still involves choices about masks, predictors and teacher-update heuristics.
Improving their accuracy--cost trade-off remains an important part of developing useful predictive representations in JEPAs.

A complementary line of work learns representations by aligning augmented views.
BYOL, SimSiam, DINO and iBOT use different forms of architectural asymmetry or teacher supervision to learn informative features~\cite{grill2020bootstrap,chen2021exploring,caron2021emerging,zhou2021ibot}.
More recently, LeJEPA combines view alignment with distributional regularisation, allowing training without an EMA teacher or stop-gradient~\cite{balestriero2025lejepa}.
This simplifies the training machinery and offers a scalable route to representation learning.
View alignment, however, does not by itself specify how to predict a hidden patch from sparse context.
That task calls for an explicit prediction mechanism, as in I-JEPA.
The challenge is to retain useful spatial information without making pre-training prohibitively expensive.

Our starting point is our earlier PhiNet framework~\cite{ishikawa2025phinets}, inspired by the temporal prediction hypothesis of the hippocampal formation~\cite{chen2024predictive}.
PhiNet uses Siamese student encoders and two successive predictors: one estimates another view's representation, and the second predicts the original image representation.
The original study reports greater robustness to collapse across weight-decay settings than SimSiam and BYOL, together with favourable online and continual-learning results for its variants.
Moreover, simplified learning-dynamics analysis offers a partial explanation.
Subsequent work includes PhiNet v2~\cite{yamada2025phinet} and SiamJEPA~\cite{yamada2026siamjepa}. However, how to realise PhiNet's predictive learning principle in a simpler architecture with stable training remains an open question.

To investigate a simpler architecture, we start from the observation that SiamJEPA's Siamese students perform on par with a single student under the same masking scheme~\cite{yamada2026siamjepa}. We adopt this single-student configuration as our baseline, Simple I-JEPA. Building on it, we introduce \textbf{\method}, which combines a sparse-context student and an EMA teacher with PhiNet's two-stage prediction principle (\cref{fig:method}).
More specifically, one student encodes sparse context from a photometrically corrupted image ($a_1$), while an EMA teacher encodes an independently corrupted view ($a_2$) and the clean image.
A single Transformer block (CA3) predicts hidden tokens of the $a_2$ view, and a token-wise MLP (CA1) maps these predictions to clean teacher tokens at the same positions.
The independent target corruption encourages prediction of content shared across views; the clean target supplies an appearance reference.
Whereas I-JEPA avoids hand-crafted augmentations~\cite{assran2023self}, \method uses photometric corruption as part of this two-stage objective.
Controlled variants (\cref{sec:design}) and direct hidden-token evaluations test the contribution of this combination.
Can the same predictions that reveal the semantic labels of hidden patches also help a planner act when an object is partly occluded?
Experiments demonstrate improved dense features and hidden-region prediction at a practical pre-training cost, with benefits for latent planning under occlusion and blur.
In particular, the hidden-token predictor, which JEPAs usually discard after pre-training, can be reused without fine-tuning to complete occluded tokens for planning.

\vspace{0.5em}
\noindent Our contributions are:
\begin{itemize}[leftmargin=*,itemsep=1pt,topsep=2pt]
\item \textbf{A two-stage latent denoising JEPA.} \method realises PhiNet's two-stage prediction with masked Transformers, with independently corrupted and clean targets at aligned hidden positions; controlled variants show that combining photometric corruption with the second stage improves dense features over the tested one-stage alternatives.
\item \textbf{Semantic information in predicted regions.} Given the same visible patches, a linear probe on \method's predicted tokens recovers hidden-region labels at 49.1 mIoU on PASCAL VOC, compared with 44.3 for Simple I-JEPA.
\item \textbf{Reusing the JEPA predictor in latent world models.} The hidden-token predictor of a JEPA is usually discarded after pre-training. We show that it can be reused, without fine-tuning, as a token-completion module for a frozen-encoder world model: under simultaneous occlusion and blur on PushT, completion with \method's predictor raises open-loop planning success from 0.33 to 0.65, whereas Simple I-JEPA's predictor reaches 0.25.
\end{itemize}

%% file: sec/3_method_figure.tex
\begin{figure*}[t]
  \centering
  \includegraphics[width=\linewidth]{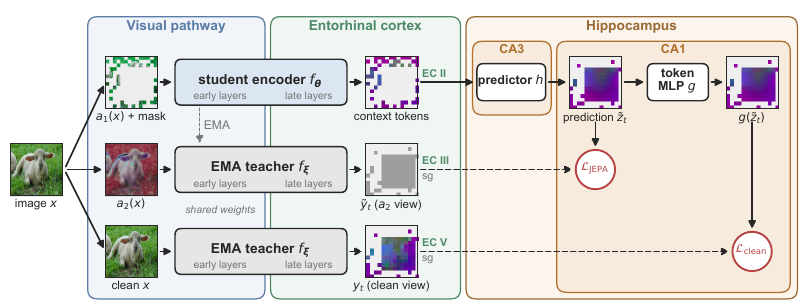}
  \caption{\textbf{Two-stage masked prediction.}
  The student encodes a sparse context corrupted by $a_1$.
  The single Transformer-block predictor $h$ (CA3) estimates hidden tokens of a second view corrupted by an independent draw $a_2$.
  The token-wise MLP predictor $g$ (CA1) maps these estimates to clean teacher tokens at the same locations.
  Both teacher branches share the EMA encoder and supply corrupted-view and clean-view targets. Gradients update the student, $h$ and $g$.
  The hippocampal correspondence inherited from PhiNet is detailed in \cref{sec:suppl-phinet}.
  sg denotes stop-gradient.
  Images show one VOC example with illustrative corruptions. Token maps use PCA colours of the default model's tokens at the evaluation context and target positions; the $a_2$ targets are schematic.}
  \label{fig:method}
\end{figure*}

%% file: sec/2_related.tex
\section{Related Work}
\label{sec:related}

\paragraph{Masked Modeling.}
Masked modelling learns representations by predicting missing content from partial observations.
MAE~\cite{he2022masked} reconstructs masked pixels with an asymmetric encoder--decoder, and SimMIM~\cite{xie2022simmim} uses a simple pixel-reconstruction head.
BEiT~\cite{bao2022beit} predicts discrete visual tokens.
CAE~\cite{chen2023context} separates latent prediction from reconstruction, while data2vec~\cite{baevski2022data2vec,baevski2023efficient} predicts contextualised teacher representations across modalities.
Denoising autoencoders~\cite{vincent2008extracting} similarly learn by reconstructing clean inputs from corruption.
\method uses sparse visible patches but supervises both prediction stages in teacher-feature space.

\paragraph{Joint Embedding Architecture.}
Joint embedding methods learn by aligning representations of augmented views of the same image.
SimCLR~\cite{chen2020simple} uses a contrastive objective, while BYOL~\cite{grill2020bootstrap} and SimSiam~\cite{chen2021exploring} use an asymmetric predictor with stop-gradient (and, for BYOL, an EMA target). %
DINO~\cite{caron2021emerging} and iBOT~\cite{zhou2021ibot} learn through self-distillation, with iBOT also supervising patch tokens.
LeJEPA~\cite{balestriero2025lejepa} combines view alignment with SIGReg, enabling training without an EMA teacher or stop-gradient.
These approaches learn features through view alignment; \method explicitly predicts hidden spatial content.

\paragraph{Joint Embedding Predictive Architecture.}
JEPAs predict target representations from context~\cite{path}.
I-JEPA~\cite{assran2023self} predicts hidden image-region features, and V-JEPA~\cite{bardes2024revisiting} extends masked latent prediction to video.
DSeq-JEPA~\cite{he2026dseq} orders target prediction by saliency.
N-JEPA~\cite{qiu2025improving} perturbs masked-token positional embeddings and trains parallel noisy and noise-free predictors to match both teacher features and one another.
StoP~\cite{bar2024stochastic} adds noise to the positions of the masked tokens, and C-JEPA~\cite{mo2024connecting} adds a VICReg regulariser to I-JEPA.
IWM~\cite{garrido2024learning} conditions its JEPA predictor on photometric transformation parameters to predict their effects in latent space.
\method receives no corruption parameters and instead uses a second, token-wise prediction stage with a clean target.
D-JEPA~\cite{chen2025denoising} combines JEPA feature prediction with a generative diffusion or flow-matching loss on image tokens, using raw pixels or VAE latents.
\method instead uses sequential predictors supervised by independently corrupted and clean teacher features, without an image-generation loss.
For world modelling, V-JEPA~2~\cite{assran2025v} and LeWorldModel~\cite{maes2026leworldmodel} learn action-conditioned latent prediction, while DINO-WM~\cite{zhou2025dino} trains dynamics on frozen DINOv2~\cite{oquab2024dinov2} features.
We use DINO-WM to compare frozen encoders for planning with clean, occluded and photometrically degraded observations.

\paragraph{World models under partial observation.}
Control from images is a partially observable problem~\cite{kaelbling1998planning}: a single frame may not reveal the full state.
Latent world models such as World Models~\cite{ha2018recurrent}, PlaNet~\cite{hafner2019learning} and DreamerV3~\cite{hafner2025mastering} handle this with a recurrent belief state that integrates observations over time.
DINO-WM instead conditions on a short history of frozen-encoder tokens without an explicit belief state.
The encoded occluded regions therefore enter the dynamics model and planning cost directly.
A frozen I-JEPA predictor can also complete the occluded features of other encoders for classification~\cite{nguyen2026jepa}; we use the encoder's own predictor to complete occluded observations of a world model.
We insert occluders at test time to assess whether pre-training on hidden-content prediction improves state representations and planning under partial observation.

\paragraph{Hippocampus-inspired self-supervised learning.}
PhiNet~\cite{ishikawa2025phinets} introduced hippocampus-inspired self-supervised predictive learning, using successive CA3--CA1 predictors motivated by the hippocampal temporal prediction hypothesis~\cite{chen2024predictive}. Its X-PhiNet variant introduces momentum-encoder targets.
Later, DeKraker et al.~\cite{dekraker2026understanding} proposed, from a neuroscience perspective, that the hippocampus acts as a self-supervised predictive learning engine at the apex of the cortical hierarchy, generating latent training signals that support cortical learning; this view is consistent with the motivation of PhiNet.
Building on PhiNet, subsequent work learns from image sequences with a variational objective~\cite{yamada2025phinet}, and SiamJEPA~\cite{yamada2026siamjepa} brings PhiNet's Siamese students to masked JEPAs.
A central finding of SiamJEPA, however, is that a single student performs on par with the Siamese configuration under the same masking scheme. This finding motivates Simple I-JEPA, the single-student baseline on which we build.
Our contribution applies PhiNet's two-stage prediction principle to this simpler configuration, extending it from convolutional, image-level learning to masked Transformer tokens.
A single Transformer block predicts hidden tokens of an independently corrupted teacher view, and a token-wise MLP then predicts the corresponding clean teacher features.
We evaluate this formulation through controlled objective ablations and direct hidden-region prediction tests.

%% file: sec/3_method.tex
\section{PhiJEPA}
\label{sec:method}

\method instantiates PhiNet's two-stage prediction principle~\cite{ishikawa2025phinets} in a masked Transformer.
The design replaces PhiNet's Siamese student configuration with a sparse-context student and two teacher branches for spatially aligned hidden-token prediction.
The first predictor is a single Transformer block; the second is a token-wise MLP.
We first describe Simple I-JEPA, which uses SiamJEPA's masking and training recipe~\cite{yamada2026siamjepa} with a single student encoder, then introduce the two-stage objective.

\subsection{Simple I-JEPA}
\label{sec:simple}
\paragraph{Input and masks.}
Let $\vx$ denote an image after a random resized crop and horizontal flip; we call it the clean image, since \method later adds photometric corruptions to it.
The image, of $H\times W$ pixels, is divided into $N=(H/P)(W/P)$ non-overlapping patches of $P\times P$ pixels.
For each image we sample a square target block $\mathcal B$ of $b\times b$ patches.
From the $N-b^2$ patches outside the block, we sample a context set $\mathcal C$ of $n_c$ patches and a set $\mathcal S$ of $n_s$ scattered targets; the remaining $N-b^2-n_c-n_s$ patches are neither seen nor predicted.
The target set is $\mathcal T=\mathcal B\cup\mathcal S$, with $|\mathcal T|=b^2+n_s$.
The block requires prediction across a large hidden region, while the scattered targets provide predictions near the context.
In this paper, $H=W=224$ and $P=16$, so that $N=196$ on a $14\times14$ grid; we use $b=9$, $n_c=49$ and $n_s=17$, so the student sees 25\% of the patches, $|\mathcal T|=98$, and 49 patches are unused.
Three properties of this mask are important (\cref{tab:masking}).
First, encoding only a quarter of the patches keeps the student cost low.
Second, the contiguous target block matters: replacing it with the same number of randomly scattered targets lowers ImageNet linear accuracy from 72.1\% to 50.2\%.
Third, predicting every hidden patch is unnecessary in this setting: using all 147 non-context patches as targets lowers accuracy to 69.5\%.

\paragraph{Encoders and tokens.}
The student encoder $f_{\vtheta}$ is a ViT that receives only the context patches $\vx_{\mathcal C}$.
It outputs one $D$-dimensional \emph{context token} per patch, $f_{\vtheta}(\vx_{\mathcal C})\in\mathbb R^{n_c\times D}$.
A teacher encoder $f_{\vxi}$ with the same architecture receives the full image and outputs one token for each of the $N$ patches; we write $\vy_t=f_{\vxi}(\vx)_t$ for its token at position $t$.
The teacher is not trained by gradients: its parameters are an exponential moving average (EMA) of the student's, $\vxi\leftarrow\tau\vxi+(1-\tau)\vtheta$.

\paragraph{Predictor and objective.}
A predictor $h$ maps the context tokens to predictions at the target positions.
It projects the context tokens to its width and places a mask token with a positional embedding at each of the $N-n_c$ non-context positions.
A single predictor block followed by LayerNorm lets each mask token attend to the context tokens and the other mask tokens and predicts all targets in one pass; only the target positions enter the loss.
We write $\vz_t=h(f_{\vtheta}(\vx_{\mathcal C}))_t$ for the prediction at target $t\in\mathcal T$, read from its mask token.
Simple I-JEPA trains the student and the predictor so that each prediction matches the teacher's token at the same position:
\begin{equation}
\mathcal L_{\rm JEPA}=\frac{1}{|\mathcal T|}\sum_{t\in\mathcal T}d(\vz_t,\sg(\vy_t)),\quad
 d(\vu,\vv)=2-2\frac{\vu^\top\vv}{\|\vu\|\|\vv\|},
\label{eq:jepa}
\end{equation}
where $\sg$ denotes stop-gradient.
The complete recipe and its comparison with I-JEPA are given in \cref{sec:suppl-simple,tab:recipe}.

\subsection{Independently corrupted and clean targets}
\label{sec:denoising}
\method builds directly on Simple I-JEPA (\cref{fig:method}).
It retains the masks $\mathcal C$ and $\mathcal T$, student $f_{\vtheta}$, EMA teacher $f_{\vxi}$ and first predictor $h$.
It changes the input views and adds a second predictor, with a separate teacher target for each stage.

\paragraph{Corrupted views.}
A photometric corruption $a$ is drawn from a distribution $\mathcal A$: with probability $p$ the image is corrupted by colour jitter, grayscale conversion, Gaussian blur, solarisation and additive Gaussian noise, each applied with its own probability (\cref{sec:suppl-dejepa}), and otherwise it is left unchanged.
These operations change pixel values but not their positions, so the same index sets $\mathcal C$ and $\mathcal T$ refer to the same image regions in every view.
For each image we draw two corruptions independently, $a_1,a_2\sim\mathcal A$.
The student receives the context patches of $\widetilde{\vx}_1=a_1(\vx)$, and the teacher encodes two full views, $\widetilde{\vx}_2=a_2(\vx)$ and the clean image $\vx$.
Unlike I-JEPA, which was designed to learn without hand-crafted augmentations~\cite{assran2023self}, \method uses photometric corruption together with two-stage supervision.
Without corruption ($p=0$), both stages target the same clean teacher tokens; the additional clean-target term remains unless $\lambda=0$.
With corruption and the two-stage objective, VOC segmentation improves from 66.2 to 69.2 mIoU (\cref{tab:ablation}).

\paragraph{First stage: predicting an independently corrupted view (CA3).}
The first predictor $h$ predicts each hidden position from the corrupted context.
Its target is the teacher token of the independently corrupted view,
\begin{equation}
\widetilde{\vz}_t=h(f_{\vtheta}(\widetilde{\vx}_{1,\mathcal C}))_t,\qquad
\widetilde{\vy}_t=f_{\vxi}(\widetilde{\vx}_2)_t .
\end{equation}
Independently sampling the target corruption changes the prediction task.
If the target were the student's own corrupted view $a_1(\vx)$, the predictor could carry the corruption it sees in the context over to its prediction.
If it were the clean image, a single predictor would have to infer the hidden content and undo the corruption at the same time.
Since the sampled target corruption $a_2$ cannot be recovered from the input corruption $a_1$, this objective encourages prediction of content shared across corrupted views.

\paragraph{Second stage: mapping to the clean target (CA1).}
A token-wise MLP $g$ then maps each first-stage prediction to the clean teacher token at the same position, $\vy_t=f_{\vxi}(\vx)_t$.
The MLP has two layers with a GELU activation and no attention.
It acts on each target token independently, mapping the first-stage prediction to the clean representation without gathering additional context.
Gradients from this loss pass through $h$ to the student, allowing the clean target to shape the learned representation.

\paragraph{Objective.}
The two stages are trained jointly,
\begin{equation}
\mathcal L=\frac{1}{|\mathcal T|}\sum_{t\in\mathcal T}
\Big[\underbrace{d(\widetilde{\vz}_t,\sg(\widetilde{\vy}_t))}_{\mathcal L_{\rm JEPA}}+\lambda\,\underbrace{d(g(\widetilde{\vz}_t),\sg(\vy_t))}_{\mathcal L_{\rm clean}}\Big],
\label{eq:dejepa}
\end{equation}
with the distance $d$ of \cref{eq:jepa}; averaged over the targets, the two terms are $\mathcal L_{\rm JEPA}$, the objective of \cref{eq:jepa} with the targets of the independently corrupted view, and $\mathcal L_{\rm clean}$, which supervises the CA1 output with the clean targets.
Both teacher views use the same EMA encoder $f_{\vxi}$, without masking.
We use $p=1$, $\lambda=0.5$ and an MLP of widths $768\to2048\to768$; setting $p=0$ and $\lambda=0$ recovers Simple I-JEPA (\cref{eq:jepa}).

\paragraph{Use after training.}
For frozen-feature evaluations and as the encoder of world models we use the EMA teacher.
To complete hidden tokens, we keep the student encoder and the first predictor and report its outputs unless stated otherwise.

\subsection{\texorpdfstring{Design ablations}{Design ablations}}
\label{sec:design}
\label{sec:exp-ablation}
The two stages of \method follow PhiNet~\cite{ishikawa2025phinets}: a first predictor that estimates another view's representation (CA3) and a second that maps it to the representation of the original input (CA1); the correspondence with the hippocampal formation is detailed in \cref{sec:suppl-phinet}.
We compare several simple extensions of Simple I-JEPA: direct denoising to a clean target, prediction of a corrupted view, and two-stage prediction with different target views or supervision positions.
All variants use the same pre-training setup (ViT-B/16, 400 epochs, $p=0.5$; \cref{tab:ablation}).
The comparisons address four design questions.
Each variant is evaluated by linear VOC segmentation of its frozen patch features and by an attentive ImageNet probe (protocols in \cref{sec:setup,sec:suppl-protocols}).

\begin{table}[t]
\centering
\small
\setlength{\tabcolsep}{2.5pt}
\begin{adjustbox}{max width=\linewidth}
\begin{tabular}{@{}llcccc@{}}
\toprule
& student & CA3 target & CA1 & VOC & Attn. \\
\midrule
(a) Simple I-JEPA & clean & clean & -- & 66.2 & 73.7 \\
(b) & $a_1$ & clean & -- & 65.4 & 74.3 \\
(c) & $a_1$ & $a_2$ (indep.) & -- & 65.9 & 73.8 \\
(d) & $a_1$ & $a_1$ (same) & \checkmark & 67.8 & 74.8 \\
(e) \textbf{\method} & $a_1$ & $a_2$ (indep.) & \checkmark & \best{69.2} & \best{75.8} \\
\bottomrule
\end{tabular}
\end{adjustbox}
\caption{\textbf{Ablations} (ViT-B/16, 400 epochs, corruption probability $p=0.5$). (c) is \method without the second stage. Student: input of the student; CA3 target: view encoded by the teacher for the predictor; CA1: \checkmark\ if the token-wise head $g$ maps the CA3 predictions to the clean teacher tokens at the same (hidden) positions, -- if there is no second stage. Variants (f) and (g), which apply the head at the visible positions, are in \cref{tab:ablation-context}. VOC: linear segmentation (mIoU); Attn.: ImageNet attentive probe (top-1).}
\label{tab:ablation}
\end{table}

\paragraph{Why not corrupt the input and predict the clean teacher?}
This one-stage denoising objective is the simplest way to use photometric corruption in a JEPA.
It does not improve dense features: corrupting the student and predicting clean teacher tokens (b) lowers VOC segmentation from 66.2 to 65.4 mIoU.
Predicting an independently corrupted view instead (c) does not help either (65.9 mIoU; ADE20K 28.0 \vs 29.2 for (a), not shown in the table).
Thus, corruption alone does not improve dense features in these comparisons.

\paragraph{Why a second stage?}
Adding the token-wise head that maps the first prediction to the clean teacher token turns the same corruption into a gain: from (c) to \method (e), VOC improves from 65.9 to 69.2 mIoU and ADE20K from 28.0 to 31.2, and the attentive probe from 73.8 to 75.8\%.
The second stage helps at every evaluated training checkpoint (\cref{sec:exp-dynamics}).
The design assigns different targets to the two predictors: the first estimates hidden content in another corrupted view, and the second maps it to clean teacher features.

\paragraph{Why an independently corrupted first-stage target?}
With the same second stage, predicting a view corrupted independently of the student's input (e) is better than predicting the student's own corrupted view (d), by 1.4 VOC mIoU.
When both views share the corruption, the predictor can reproduce appearance changes visible in the input.
Independent corruption removes this direct correspondence and encourages prediction of shared content.

\paragraph{What if CA1 supervises the context patches?}
We also test applying CA1 to the predictor's outputs at the 49 context-patch positions supplied to the student encoder, matching them to clean teacher tokens at the same positions, with a clean first-stage target (\cref{tab:ablation-context}). This gives 65.4 VOC mIoU with clean student inputs and 68.3 with corrupted inputs. The latter is below the full objective (69.2 mIoU), which changes both the CA1 positions and the first-stage target.

\paragraph{Summary.}
The PhiNet arrangement gives the strongest dense-feature results among the tested extensions: an independently corrupted first-stage target followed by a clean target at hidden positions (\cref{tab:ablation,tab:ablation-context}).
It costs one additional teacher pass and a small head, while the student still encodes only the sparse context.

%% file: sec/4_experiments.tex
\section{Encoder evaluation}
\label{sec:experiments}
We first evaluate the pre-trained encoder through frozen-feature quality, training cost, robustness and the semantic content of hidden-region predictions.
These evaluations precede the main world-model experiments in \cref{sec:wm}.

\subsection{Setup}
\label{sec:setup}
\paragraph{Pre-training.}
Simple I-JEPA and \method use ViT-B/16~\cite{dosovitskiy2021image} and are pre-trained on ImageNet-1K~\cite{deng2009imagenet} for 400 epochs.
Their optimiser, masking and first predictor are identical.
We use AdamW~\cite{loshchilov2019decoupled}, effective batch size 8192, a peak learning rate of $4.8\times10^{-3}$ with 40 warm-up epochs and cosine decay, and weight decay 0.1.
The default \method uses $p=1$, $\lambda=0.5$, and a $768\to2048\to768$ CA1 head.
The $p=0.5$ configuration is used for the main objective ablations.
The default was chosen using the reported evaluations.
Full augmentation, EMA and probe settings are in \cref{sec:suppl-protocols,tab:hparams}.

\paragraph{Comparisons.}
Simple I-JEPA provides the controlled comparison for the objective.
We also evaluate pixel completion with MAE in the appendix (\cref{tab:masked-pred,tab:occlusion}).
I-JEPA ViT-B/16 is our pre-training with the official code (\cref{tab:main}).
The prediction, retrieval and probe evaluations use frozen pre-trained networks, with one pre-training run per configuration.

\subsection{Frozen-feature quality and robustness}
\label{sec:exp-quality}

\paragraph{Dense semantics and image-level discrimination.}
We evaluate final-layer patch features with our common linear and attentive probes (\cref{tab:main}).
For segmentation, a BatchNorm and linear head are trained on frozen full-image features at 448 pixels; the ADE20K evaluation uses 150 classes.
These scores differ from the 224-pixel target-only evaluation and are not fine-tuned segmentation results.
With the default $p=1$, \method improves VOC and ADE20K over Simple I-JEPA by 3.1 and 1.5 mIoU, respectively ($p=0.5$, as in \cref{tab:ablation}: 3.0 and 2.0).
It also exceeds DSeq-JEPA by 5.5 and 2.4 mIoU, while DSeq-JEPA remains ahead in linear and attentive classification.
The main gains are in patch-level tasks: \method approaches iBOT on VOC (69.3 \vs 70.9) but not on ADE20K or attentive classification, and its mean-token linear accuracy barely changes from the baseline.
Probe optimisation and feature geometry can also affect these rankings.
For the $p=0.5$ checkpoint, variation across three segmentation-head seeds is $\pm0.2$ mIoU.
This measures probe variation with the encoder fixed, rather than pre-training variance.

\begin{table}[t]
\centering
\small
\setlength{\tabcolsep}{3pt}
\begin{adjustbox}{max width=\linewidth}
\begin{tabular}{@{}lcccccc@{}}
\toprule
Method & Ep. & FLOPs & Lin. & Attn. & VOC & ADE \\
\midrule
\multicolumn{7}{@{}l}{\emph{view invariance}} \\
DINO~\cite{caron2021emerging} & 400 & -- & 71.3 & 78.3 & 60.4 & 30.4 \\
iBOT~\cite{zhou2021ibot} & 400 & 263 & \best{74.2} & \best{79.9} & \best{70.9} & \best{35.6} \\
LeJEPA$^\ddagger$~\cite{balestriero2025lejepa} & 300 & 121 & 70.9 & 77.8 & 67.8 & 31.9 \\
\midrule
\multicolumn{7}{@{}l}{\emph{masked latent prediction}} \\
I-JEPA$^\S$~\cite{assran2023self} & 600 & 75 & 70.8 & 73.7 & 61.4 & 25.4 \\
DSeq-JEPA~\cite{he2026dseq} & 600 & $\approx$83 & \best{73.7} & \best{75.8} & 63.8 & 28.3 \\
Simple I-JEPA (ours) & 400 & 36 & 72.1 & 73.7 & 66.2 & 29.2 \\
\method, $p=0.5$ (\cref{tab:ablation}) & 400 & 56 & 72.4 & \best{75.8} & 69.2 & \best{31.2} \\
\method, $p=1$ (default) & 400 & 56 & 72.0 & 75.6 & \best{69.3} & 30.7 \\
\bottomrule
\end{tabular}
\end{adjustbox}
\caption{\textbf{Frozen representation quality.} All checkpoints use ViT-B/16 and our evaluation protocols. Ep./FLOPs: pre-training epochs and total training FLOPs ($\times10^{18}$), counted on one training step of each method's code ($\approx$: estimated; --: not counted); measured or estimated GPU-hours are in \cref{tab:compute}. Lin./Attn.: ImageNet-1K top-1 (\%); VOC/ADE: linear segmentation mIoU at 448 pixels. Bold: best within the view-invariance and masked-latent-prediction groups. Other methods use public checkpoints, except LeJEPA$^\ddagger$, a third-party reimplementation~\cite{ai2026lejepa}. $^\S$Re-trained with the official I-JEPA code and the same recipe as DSeq-JEPA~\cite{he2026dseq} (600 epochs), and evaluated with our protocols, which differ from those of the original I-JEPA paper (it reports 72.9\% linear-probe accuracy with its own protocol).}
\label{tab:main}
\end{table}

\begin{figure*}[t]
  \centering
  \includegraphics[width=\linewidth]{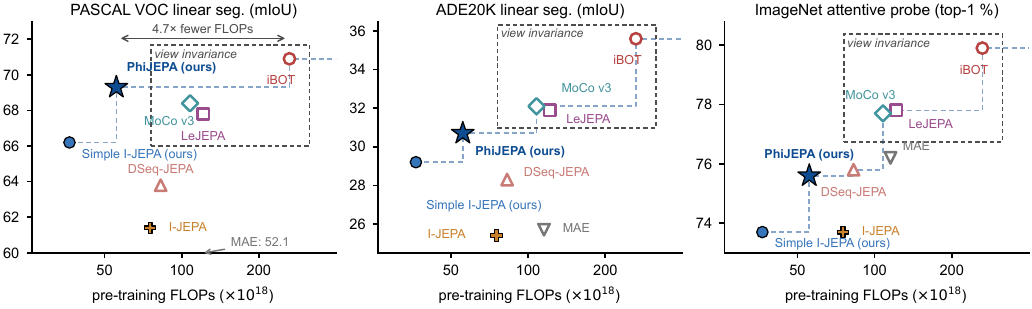} %
  \caption{\textbf{Accuracy and pre-training cost.} Evaluated ViT-B/16 checkpoints on three frozen-feature tasks; horizontal axes show total training FLOPs on a log scale (\cref{tab:compute}; the same plot with GPU-hours is \cref{fig:efficiency-gpuh}). Filled markers: checkpoints we trained (I-JEPA: our re-training, \cref{tab:main}); open markers: public checkpoints. FLOPs are counted on one training step of each method's code, with DSeq-JEPA shown as a time-ratio-based proxy whose compute ranking requires direct verification. MoCo v3 (VOC 68.4, ADE20K 32.1, Attn.\ 77.7) and MAE (52.1, 25.7, 76.2) use the same protocols and are not listed in \cref{tab:main}. The dashed front is restricted to the plotted checkpoints and these costs. \method is within 1.6 VOC mIoU of iBOT with 4.7$\times$ fewer training FLOPs. The comparisons use different training schedules and do not establish superiority under matched compute.}
\label{fig:efficiency}
\end{figure*}

\paragraph{Pre-training cost.}
We compare training FLOPs, counted using a common convention for the implemented training operations (\cref{tab:compute}); measured GPU-hours on our node are reported in the appendix (\cref{sec:suppl-cost}).
Per image, a training step of \method costs 109 GFLOPs, compared with 71 for Simple I-JEPA.
The student still encodes only 49 context patches; the overhead comes from the teacher pass on the independently corrupted view and the CA1 head.
Over 400 epochs, \method needs $56\times10^{18}$ FLOPs and Simple I-JEPA $36\times10^{18}$.
A training step of I-JEPA costs a similar 98 GFLOPs, but its 600-epoch schedule needs $75\times10^{18}$ FLOPs. The DSeq-JEPA value of $83\times10^{18}$ is a time-ratio-based proxy (\cref{sec:suppl-cost}).
\method yields higher VOC and ADE20K segmentation scores than these I-JEPA and DSeq-JEPA checkpoints with 25\% fewer training FLOPs than I-JEPA, and iBOT's schedule needs 4.7$\times$ the FLOPs of \method.
Simple I-JEPA uses fewer FLOPs and scores lower on both segmentation tasks.
These are checkpoint-level accuracy--cost comparisons, not equal-compute training experiments.
\Cref{fig:efficiency} places these measurements alongside the other evaluated checkpoints on all three tasks.

\paragraph{Appearance corruption and missing observations.}
On ImageNet, \method is markedly more robust than Simple I-JEPA to photometric corruptions (64.9\% \vs 54.3\% attentive accuracy at severity 3), but less robust to a large image occluder in classification (47.1\% \vs 50.9\%); appearance corruption and missing observations are different tests (\cref{tab:robust}, appendix).

\subsection{Semantic content of hidden-region predictions}
\label{sec:exp-predict}
We use the 1,449 PASCAL VOC validation images~\cite{everingham2009pascal} at 224 pixels.
Each model receives the same context and target mask for each image, with 25\% of patches visible and 50\% evaluated as targets.
For each encoder, a BatchNorm and linear segmentation head are trained on clean full-image teacher features from the training set, then applied to predicted target tokens.
The head learning rate is selected on clean validation teacher features.
We report mIoU over target pixels only.
This evaluates whether the predicted tokens preserve semantic information readable by a head trained in the corresponding teacher space.

\begin{table}[t]
\centering
\small
\setlength{\tabcolsep}{4pt}
\begin{tabular}{@{}lccc@{}}
\toprule
& \shortstack{Simple\\I-JEPA} & \shortstack{PhiJEPA\\$p=0.5$} & \shortstack{PhiJEPA\\$p=1$} \\
\midrule
Full-image teacher & 65.0 & 66.4 & 67.7 \\
Prediction, clean input & 44.3 & 48.5 & \best{49.1} \\
Prediction, corrupted input & 35.8 & \best{44.6} & 44.5 \\
Prediction, far targets & 43.9 & 47.6 & \best{48.4} \\
Nearest token, far targets & 32.6 & 32.4 & 32.9 \\
\bottomrule
\end{tabular}
\caption{\textbf{Semantic prediction from the same visible context.}
VOC target-pixel mIoU for ViT-B/16 models.
Far targets are at least two patches from every visible patch and account for 61\% of the target set.
The full-image teacher row uses clean input.
Corrupted input uses a fixed draw of the training corruption, with clean labels and teacher targets.
Detailed comparisons are in \cref{tab:masked-pred}.}
\label{tab:prediction-main}
\end{table}

\begin{figure}[t]
  \centering
  \includegraphics[width=0.7\linewidth]{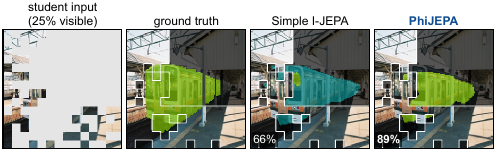}
  \caption{\textbf{Predicting a hidden region} (VOC val). Both models see the same 25\% of the patches; the predicted tokens of the target patches (outlined) are segmented by each model's linear head; numbers: pixel accuracy over the targets.}
  \label{fig:masked-pred-main}
\end{figure}

\paragraph{Accuracy of predicted content.}
\method improves hidden-region segmentation by 4.8 mIoU over Simple I-JEPA (\cref{tab:prediction-main}).
Full-image teacher accuracy improves by 2.7 points under the same evaluation.
The larger gain on predicted tokens is consistent with improved semantic completion, although the score also reflects encoder quality and compatibility with each model's probe.
Hidden-region accuracy is similar before and after the denoising head (49.1 and 48.9 mIoU, respectively), so the second stage helps mainly through training (\cref{tab:ablation}) rather than by improving this metric at inference.
On far targets, \method reaches 48.4 mIoU and exceeds nearest-visible-token copying by about 15 points.
The copying baseline uses contextualised tokens from a full-image teacher pass, so it can carry information from hidden patches.
The prediction gains persist in the interior of the hidden block.
In \cref{fig:masked-pred-main}, Simple I-JEPA labels the hidden train as a bus, whereas \method predicts the correct class from the same visible patches.
Further examples, including a failure case, are in \cref{fig:masked-pred}.

\paragraph{Corrupted inputs and visible context.}
Corrupting the visible input reduces \method's mIoU by 4.6 points, compared with 8.5 for Simple I-JEPA.
The $p=0.5$ variant gives similar results on this mask distribution.

%% file: sec/5_worldmodel_noproxy.tex
\section{World-model experiments}
\label{sec:wm}
We now evaluate the frozen encoders within a latent world model, testing planning under occlusion and photometric corruption and measuring state retrieval from occluded frames.

\subsection{Setup}
\label{sec:wm-setup}
We compare six frozen encoders within DINO-WM~\cite{zhou2025dino} on PushT: \method (default $p=1$), Simple I-JEPA, DINOv2-B, LeJEPA~\cite{balestriero2025lejepa}, DSeq-JEPA~\cite{he2026dseq} and iBOT~\cite{zhou2021ibot}. All encoders are general-purpose, with no pre-training or fine-tuning on PushT. DINOv2-B was pre-trained on LVD-142M~\cite{oquab2024dinov2}, while the others use ImageNet-1K~\cite{deng2009imagenet}. DINOv2-B is distilled from a ViT-g and uses 14-pixel patches; as in DINO-WM, its input is resized from 224 to 196 pixels, so it also produces a $14\times14$ token grid. DINOv2 is the encoder family used by DINO-WM (ViT-S/14 there); we use DINOv2-B to match the size of the other encoders, and use ImageNet-1K-pre-trained iBOT ViT-B/16 for a comparison matched in architecture and pre-training dataset. For each encoder, we train a world model with the same training schedule and batch size, following DINO-WM's frame skip of 5 and three history frames; unlike its default recipe, we train without a decoder at batch size 128, with bfloat16 precision on two H100 GPUs. Encoder tokens are scaled to unit RMS.

To evaluate partial observations, we cover cells on the $14\times14$ observation grid with a magenta occluder and keep the goal image clean.
Because every encoder uses a $14\times14$ token grid, each occluded cell corresponds to one token; completion replaces the tokens of the cells whose pixels are mostly occluder-coloured. We compare the encoded occluded observations with completion using nearest visible tokens, MAE pixel inpainting or the encoder's own predictor, where available (\cref{tab:planning,tab:corruption}). \method and Simple I-JEPA provide separate hidden-token predictors; for \method, planning uses the first predictor followed by the CA1 head, whose output is trained to match the clean teacher tokens. LeJEPA has no such predictor, the public DSeq-JEPA checkpoint contains only the encoder, and iBOT predicts masked tokens through its encoder.

For planning, we use the cross-entropy method (CEM) with a horizon of 5, 300 samples and 30 elites per iteration, and 30 optimisation iterations. Each episode executes one open-loop plan, following the CEM configuration of DINO-WM's official code. Published DINO-WM headline results use model predictive control with replanning. Success follows DINO-WM's criterion: joint agent-and-block position error below 20 units in PushT environment coordinates and block angle error within $20^\circ$.

\subsection{Planning with complete and occluded observations}
\label{sec:exp-planning}
\begin{table}[t]
\centering
\small
\setlength{\tabcolsep}{3pt}
\begin{adjustbox}{max width=\linewidth}

\begin{tabular}{@{}lccccc@{}}
\toprule
start observation & DINOv2-B & LeJEPA & DSeq-JEPA & iBOT & \method \\
\midrule
clean & 0.67 & 0.86 & 0.82 & \best{0.88} & 0.80 \\
\midrule
T block occluded, no completion & 0.28 & 0.29 & 0.28 & 0.18 & \best{0.33} \\
\quad + nearest visible token & 0.31 & 0.24 & 0.51 & 0.11 & \best{0.55} \\
\quad + MAE pixel inpainting & 0.26 & 0.45 & \best{0.52} & 0.43 & -- \\
\quad + own predictor & -- & -- & -- & -- & \best{0.59} \\
\bottomrule
\end{tabular}
\end{adjustbox}
\caption{\textbf{CEM planning success with complete and occluded observations} (PushT). DINO-WM models after 15 epochs with the same training schedule and batch size; mean over three planning seeds of 50 episodes each, binomial standard errors at most 0.04. Own predictor completes the occluded tokens with \method's pre-trained predictor (first predictor and CA1 head). Bold marks the highest success rate in each row. LeJEPA has no hidden-token predictor, the public DSeq-JEPA checkpoint contains only the encoder, and iBOT has no separate predictor.}
\label{tab:planning}
\end{table}

With complete observations, iBOT and LeJEPA give the highest planning success under the same world-model training recipe and budget (\cref{tab:planning}). \method achieves 0.80 success, below these two encoders and close to DSeq-JEPA. Occluding the T block substantially reduces success for every encoder, including those that perform best on clean images. This motivates evaluating whether completing the hidden regions can recover useful information for planning.

The benefit of completion depends on the encoder and the completion method. For \method, copying the nearest visible token raises success from 0.33 to 0.55, and its pre-trained predictor further raises it to 0.59. Under occlusion alone, Simple I-JEPA's predictor reaches a comparable, slightly higher 0.63 (\cref{tab:corruption}); the advantage of \method's predictor appears when the visible context is also blurred (\cref{sec:exp-corruption}). MAE pixel inpainting helps LeJEPA, iBOT and DSeq-JEPA, while DINOv2-B benefits little from either tested completion method. DSeq-JEPA gives the strongest result among the other encoders with generic completion (0.52 with MAE), approaching \method's result with nearest-token copying. Thus, \method's representation supports useful completion even through simple copying, with a further gain from its learned predictor. These comparisons use synthetic occlusion in PushT and three planning seeds per condition. A complementary state-retrieval evaluation also shows improved recovery of the occluded object's state with \method's predictor (\cref{sec:exp-occlusion}, appendix).

\subsection{\texorpdfstring{Occlusion and photometric corruption}{Occlusion and photometric corruption}}
\label{sec:exp-corruption}
To test planning when observations are both incomplete and degraded, we vary occlusion and Gaussian blur independently on PushT. We include Simple I-JEPA to assess the effect of \method's two-stage pre-training under the same world-model training budget. All world models are evaluated after 15 epochs using the CEM planner of \cref{sec:wm-setup}. Blur is applied to the start observation at $\sigma=2$ or 4 pixels, with the latter exceeding the range used in \method's pre-training. In the combined conditions, we blur the frame before adding the occluder. The goal image remains clean.

\begin{table}[t]
\centering
\small

\setlength{\tabcolsep}{3pt}
\begin{adjustbox}{max width=\linewidth}
\begin{tabular}{@{}lccccccccc@{}}
\toprule
& \multicolumn{6}{c}{no completion} & MAE & \multicolumn{2}{c}{own predictor} \\
\cmidrule(lr){2-7}\cmidrule(lr){8-8}\cmidrule(lr){9-10}
start observation & DINOv2-B & LeJEPA & DSeq & iBOT & Simple & \method & iBOT & Simple & \method \\
\midrule
clean & 0.67 & 0.86 & 0.82 & \best{0.88} & 0.77 & 0.80 & -- & -- & -- \\
occluder on the T block & 0.28 & 0.29 & 0.28 & 0.18 & \best{0.33} & \best{0.33} & 0.43 & \best{0.63} & 0.59 \\
\midrule
blur, $\sigma=2$ & 0.70 & 0.57 & 0.65 & 0.75 & 0.49 & \best{0.81} & -- & -- & -- \\
blur, $\sigma=4$ & 0.53 & 0.03 & 0.18 & \best{0.54} & 0.06 & 0.40 & -- & -- & -- \\
\midrule
blur $\sigma=2$ + occluder & 0.22 & 0.12 & 0.07 & 0.20 & 0.11 & \best{0.33} & 0.40 & 0.25 & \best{0.65} \\
blur $\sigma=4$ + occluder & 0.10 & 0.01 & 0.03 & \best{0.16} & 0.02 & 0.13 & \best{0.38} & 0.03 & 0.33 \\
\bottomrule
\end{tabular}
\end{adjustbox}
\caption{\textbf{CEM planning success under occlusion and blur} (PushT; world models after 15 epochs; mean over three planning seeds of 50 episodes each; binomial standard errors at most 0.04). Simple: Simple I-JEPA. DSeq: DSeq-JEPA (its public checkpoint has no predictor). Bold marks the best encoder without completion and the best completion. MAE: iBOT with MAE pixel inpainting of the occluded region (as in \cref{tab:planning}).}
\label{tab:corruption}
\end{table}

Blur changes which representations support effective planning (\cref{tab:corruption}). At $\sigma=2$, within the range used in \method's pre-training, \method achieves the highest success rate (0.81). LeJEPA and iBOT, which lead with complete clean observations, both lose accuracy under this perturbation. With stronger blur ($\sigma=4$), iBOT and DINOv2-B perform best, while \method remains substantially more effective than Simple I-JEPA. The ranking therefore depends on the severity of the degradation.

With both occlusion and moderate blur ($\sigma=2$), the same frozen \method encoder yields 33\% planning success when occluded tokens are left uncompleted and 65\% when its pre-trained predictor fills them in.
These results suggest that \method's two-stage pre-training helps its predictor recover useful hidden-state information even when the visible context is degraded. This benefit is strongest within the blur range used during pre-training. Note, however, that under stronger blur combined with occlusion ($\sigma=4$), iBOT with MAE inpainting performs better (0.38 \vs 0.33), indicating that the benefit depends on the severity of the degradation.

%% file: sec/5_conclusion.tex
\section{Discussion and Limitations}
\label{sec:discussion}
\paragraph{What makes \method work?}
The results support \method's two-stage prediction of hidden Transformer tokens, adopted from PhiNet.
Independent corruption alone and one-stage denoising both fall short of the full objective.
CA3 and CA1 give similar hidden-region accuracy (49.1 and 48.9 mIoU), consistent with a training role for the second stage.
The clean-input head control uses context positions (\cref{tab:ablation-context}), so these experiments do not establish that every component is universally necessary.
A matched auxiliary head with a different target would help separate the benefit of clean supervision from that of extra parameters.
Neither the hippocampal analogy nor these experiments guarantees that the Transformer avoids collapse.

\paragraph{When should \method help?}
The benefits of an objective depend on the task. %
\method is motivated by two assumptions: visible context contains information about hidden content, and photometric changes are nuisances rather than task-relevant signals.
Its strongest results are in hidden-region semantics, photometric robustness and planning under combined occlusion and blur.
It is less competitive than view-invariance methods on colour- and texture-dominated transfer (\cref{tab:transfer}), and below Simple I-JEPA on ImageNet~1\% and classification with a large occluder (\cref{tab:transfer,tab:robust}).
We expect smaller benefits when fine colour or texture differences matter, or when visible context cannot resolve the hidden state.
In the latter case, a predictor may produce plausible but incorrect content.
Deployment would also require identifying which observations are unreliable.

\paragraph{Scope of the world-model evidence.}
The world-model experiments use one environment, PushT, with synthetic occluders that leave the physics unchanged.
Our models use a shorter schedule than published DINO-WM training, with the same training schedule and batch size for the compared encoders.
The planning results (\cref{sec:wm}) use three planning seeds per condition and a single open-loop plan per episode. Predictor completion assumes that the occluded locations are known. Binomial standard errors describe episode-level uncertainty and do not quantify variation across independently trained world models.
Further environments, natural occluders that move during execution, and closed-loop replanning are needed to establish how far the robustness extends.
PushT retrieval additionally depends on the candidate-frame coverage, and nearby frames from the same trajectory can affect its accuracy.
Ambiguous hidden states further motivate predicting uncertainty alongside a completed representation.
The corruption study (\cref{sec:exp-corruption}) uses synthetic pixel-level degradations. Evaluation on real robots with natural blur, lighting changes and self-occlusion remains open.

\paragraph{Experimental limits.}
The controlled pre-training comparisons use ViT-B/16 with one seed per configuration; probe seeds do not measure pre-training variance.
The default $p$ was selected using the reported evaluations, and the main ablations use $p=0.5$.
The single-run ViT-L results (\cref{sec:suppl-vitl}) motivate broader scaling evaluations and repeated pre-training runs.

\section{Conclusion}
\label{sec:conclusion}
We introduced \method, which combines a single sparse-context student and an EMA teacher with PhiNet's two-stage prediction principle. Predicting independently corrupted teacher features and then their clean counterparts improves dense representations over the tested one-stage objectives. The resulting encoder improves hidden-region semantics and classification under photometric corruption relative to Simple I-JEPA, and achieves higher linear segmentation accuracy than the evaluated I-JEPA and DSeq-JEPA checkpoints with 25\% fewer training FLOPs than I-JEPA.
In PushT, the frozen encoder and its pre-trained predictor support latent planning from occluded observations. The benefit is particularly evident when the visible context is also blurred within the pre-training range, while with stronger blur and occlusion, iBOT with MAE inpainting performs better. On PushT, the same predictor that learned to infer missing tokens during pre-training can supply a planner with information about an occluded object, provided its occluded patches are identified at test time.

%% file: sec/6_llm_statement.tex
\subsection*{Acknowledgements}
The computations in this work were carried out using the supercomputer Genkai at the Research Institute for Information Technology, Kyushu University.
Makoto Yamada was partly supported by JSPS KAKENHI Grant Number 24K03004 and by JST ASPIRE JPMJAP2302.

\subsection*{AI use statement}
We used AI assistants extensively throughout this work.
Claude Code (Anthropic) was used for implementation and debugging, for launching, monitoring and analysing experiments (including hyperparameter choices and ablations), for literature searches, for figure preparation, and for drafting and revising the manuscript.
Codex (OpenAI) was used to review and revise the manuscript's structure, wording and \LaTeX, and to check its internal consistency.
The research direction, the method and the final decisions were made by the authors, who checked the reported numbers against the experiment logs and take full responsibility for the manuscript, code, references, experimental results and findings.

%% file: sec/X_suppl.tex
\clearpage
\appendix
\onecolumn
\begin{center}\Large\textbf{Appendix}\end{center}
\vspace{0.5em}

\section{Simple I-JEPA: details}
\label{sec:suppl-simple}
Simple I-JEPA uses SiamJEPA's masking and training recipe~\cite{yamada2026siamjepa} with a single student encoder and no second view.

\begin{table}[h]
\centering
\small
\setlength{\tabcolsep}{3pt}
\begin{adjustbox}{max width=\linewidth}
\begin{tabular}{@{}lll@{}}
\toprule
& I-JEPA~\cite{assran2023self} & Simple I-JEPA \\
\midrule
context & block (85--100\%) $\setminus$ targets & random 25\% \\
targets & 4 blocks (15--20\% each) & square + scattered (50\%) \\
predictions / image & 4 & 1 \\
predictor & 6 layers (ViT-B), width 384 & 1 block, width 768 \\
loss & squared $\ell_2$ & cosine distance \\
EMA momentum & 0.996 $\to$ 1 (linear) & 0.99 / 0.999 / 0.9999 (steps) \\
weight decay & 0.04 $\to$ 0.4 (linear) & 0.1 \\
batch size & 2048 & 8192 \\
learning rate & $10^{-3}$, 15 warm-up ep. & $4.8\times10^{-3}$, 40 warm-up ep. \\
epochs (ViT-B) & 600 & 400 \\
\bottomrule
\end{tabular}
\end{adjustbox}
\caption{\textbf{I-JEPA \vs Simple I-JEPA} (ViT-B/16).}
\label{tab:recipe}
\end{table}

\begin{table}[h]
\centering
\small
\begin{adjustbox}{max width=\linewidth}
\begin{tabular}{@{}lccc@{}}
\toprule
Masking & context & targets & Top-1 \\
\midrule
square + random (default) & 25\% & 50\% & 72.1 \\
random (no square) & 25\% & 50\% & 50.2 \\
square + random, all non-context targets & 25\% & 75\% & 69.5 \\
\bottomrule
\end{tabular}
\end{adjustbox}
\caption{\textbf{Masking strategy} for Simple I-JEPA (ViT-B/16, 400 epochs, linear probe, MAE protocol, one run each). Without the square, the 98 targets are drawn uniformly at random.}
\label{tab:masking}
\label{tab:ablation-mask}
\end{table}

Removing the square is the most harmful change: with the same 49 context and 98 target patches drawn uniformly at random, the accuracy drops from 72.1\% to 50.2\%.
Scattered targets can be predicted by interpolating nearby context patches, which agrees with I-JEPA's random-masking ablation~\cite{assran2023self}.
Predicting all 147 non-context patches reduces accuracy by 2.6 points.
These additional targets are scattered near the visible context and can often be predicted through local interpolation.
\begin{table}[h]
\centering
\small
\setlength{\tabcolsep}{4pt}
\begin{adjustbox}{max width=\linewidth}
\begin{tabular}{@{}ll@{}}
\toprule
Config & ViT-B/16 \\
\midrule
optimizer & AdamW, $\beta=(0.9,0.95)$ \\
base / actual learning rate & $1.5\times10^{-4}$ / $4.8\times10^{-3}$ \\
gradient clipping & 3.0 \\
weight decay & 0.1 \\
effective batch & 8192 ($512\times4$ GPUs $\times4$ accum.) \\
schedule & 40 warm-up epochs, cosine, 400 epochs \\
EMA momentum & 0.99 / 0.999 / 0.9999 (switch at 50\%, 75\%) \\
masking & context 49, square $9\times9$, targets 98 \\
predictor & 1 block, width 768, 16 heads \\
loss & cosine distance \\
base augmentation & random resized crop (0.2--1), flip \\
\method only & corruptions with $p=1$, CA1 MLP, $\lambda=0.5$ \\
precision & bf16 (predictor and loss in fp32) \\
hardware & 1 node, 4$\times$ NVIDIA H100 \\
\bottomrule
\end{tabular}
\end{adjustbox}
\caption{\textbf{Pre-training hyper-parameters}, shared by Simple I-JEPA and \method.}
\label{tab:hparams}
\end{table}

\section{\method: details}
\label{sec:suppl-dejepa}

\paragraph{Corruptions.}
Corruptions are sampled independently for each image.
An image enters the pipeline with probability $p$ ($p=1$ by default and $p=0.5$ in the ablations).
The following operations are then applied in order, each with its own probability: colour jitter (0.8; brightness and contrast factors in $[0.6,1.4]$, saturation factor in $[0.8,1.2]$), grayscale (0.2), separable Gaussian blur with a 23-tap kernel (0.5, $\sigma\sim\mathcal U[0.1,2]$), solarisation at 0.5 of the intensity range (0.2) and additive Gaussian noise (0.2, standard deviation 0.05 in $[0,1]$ intensity units).
All operations act on pixel values only, so the patch grid and the masks are shared by all views.
\paragraph{Teacher passes.}
The two teacher branches share the EMA encoder: it encodes $\widetilde{\vx}_2$ (target of CA3) and $\vx$ (target of CA1) without masking; when $a_2$ is the identity, the two passes coincide.

\paragraph{Variants of \cref{tab:ablation}.}
(b) is \method without CA1 and with the clean image as the CA3 target.
(f) and (g) apply the CA1 head at the 49 visible (context) positions, where the predictor's output is aligned to the clean teacher at the same positions; the CA3 target is the clean image (\cref{tab:ablation-context}).
(c) is (e) without the CA1 head: the CA3 target is the independently corrupted view $a_2(\vx)$ and there is no second stage.
(d) uses the student's own corrupted view as the CA3 target.

\begin{table}[h]
\centering
\small
\setlength{\tabcolsep}{4pt}
\begin{tabular}{@{}lccccc@{}}
\toprule
& student & CA3 target & CA1 & VOC & Attn. \\
\midrule
(a) & clean & clean & -- & 66.2 & 73.7 \\
(f) & clean & clean & \checkmark\ (visible) & 65.4 & 73.4 \\
(g) & $a_1$ & clean & \checkmark\ (visible) & 68.3 & 75.4 \\
\bottomrule
\end{tabular}
\caption{\textbf{Second stage at the visible positions} (as \cref{tab:ablation}; $p=0.5$). \checkmark\ (visible): the CA1 head maps the predictor's outputs at the 49 visible (context) positions to the clean teacher's tokens at the same positions, and the CA3 target is the clean image. Without corruption (f) the second stage does not help; with a corrupted student (g) it does. (g) and \method (e) differ both in the positions of the second stage and in the CA3 target, so they do not isolate the effect of the positions.}
\label{tab:ablation-context}
\end{table}

\section{Relation to PhiNet and the hippocampal formation}
\label{sec:suppl-phinet}
This section describes the structural correspondence between \method, PhiNet and the hippocampal temporal prediction hypothesis that motivated the two-stage design.

\method extends the two-stage prediction principle of PhiNet to masked tokens, continuing the same research direction.
PhiNet~\cite{ishikawa2025phinets} is a non-contrastive model derived from the temporal prediction hypothesis of the hippocampal formation~\cite{chen2024predictive}.
Under this hypothesis, entorhinal cortex (EC) layer~II sends its output to CA3, which predicts the next input.
CA1 compares this prediction with a direct input from EC layer~III and sends its output to the deep EC layers (layer~V).
PhiNet maps the encoder representations of two augmented views and the original image to EC layers~II, III and V.
Its first predictor corresponds to CA3.
CA1 is represented by a loss comparing the CA3 prediction with the layer~III view and a second predictor targeting the original-image representation (layer~V).
X-PhiNet computes targets with a slow EMA encoder, following the long-term-memory role in complementary learning systems~\cite{mcclelland1995why,kumaran2016what}.
\method retains this assignment at the token level (\cref{tab:phinet}).
Early encoder layers represent the visual pathway feeding the EC, and later layers represent the EC.
CA3 becomes the masked-token predictor.
The layer~III and layer~V signals are teacher tokens at hidden positions from the independently corrupted and clean views, respectively; the CA1 predictor becomes a token-wise MLP.

\begin{table}[h]
\centering
\small
\setlength{\tabcolsep}{2.5pt}
\begin{adjustbox}{max width=\linewidth}
\begin{tabular}{@{}lll@{}}
\toprule
Hippocampal formation & PhiNet / X-PhiNet & \method \\
\midrule
EC layer II $\to$ CA3 & $f(\vx^{(1)})$, view 1 & $f_{\vtheta}$ on masked $a_1(\vx)$ \\
CA3 & predictor $h$ (MLP) & $h$ (one Transformer block) \\
EC layer III $\to$ CA1 & $\sg f(\vx^{(2)})$, view 2 & $\sg f_{\vxi}(a_2(\vx))$ at targets \\
CA1: compare CA3, EC~III & Sim-1 (cosine) & $\mathcal{L}_{\rm JEPA}$ (cosine) \\
CA1: second predictor & $g$ (MLP) & $g$ (token-wise MLP) \\
CA1 $\to$ EC layer V & Sim-2 \vs $\sg f(\vx)$ & $\mathcal{L}_{\rm clean}$ \vs $\sg f_{\vxi}(\vx)$ \\
slow memory & EMA encoder (X-PhiNet) & EMA teacher $f_{\vxi}$ \\
\bottomrule
\end{tabular}
\end{adjustbox}
\caption{\textbf{Correspondence} between the hippocampal formation as modelled by PhiNet~\cite{ishikawa2025phinets} and \method. $\vx^{(1)},\vx^{(2)}$: augmented views; $\vx$: original (clean) image.}
\label{tab:phinet}
\end{table}

PhiNet uses Siamese student encoders, whereas \method uses a sparse-context student with augmented-view and clean-view teacher branches.
The first predictor changes from an image-level MLP to a single Transformer block for masked-token prediction, while the second is an MLP applied independently to each predicted token.
The representations are patch tokens, both layer~III and layer~V targets come from the EMA teacher, and the views differ only by position-preserving photometric corruptions.
Interpreting the two independent corruptions as two samples of the same scene at different moments, the layer~II and layer~III inputs carry different samples, as in the temporal prediction hypothesis.
In CA3, the predictor completes the masked tokens from the visible ones, a form of pattern completion; unlike the recurrent CA3 network, it is a single feed-forward block.
This is a structural analogy rather than a biological implementation.
We omit the dentate gyrus and subiculum, use the later layers of the student and EMA networks for the EC, and train by back-propagation.
We use this assignment because it performed best among the variants in \cref{sec:exp-ablation}.

\section{Detailed experimental protocols}
\label{sec:suppl-support}
\label{sec:suppl-protocols}

\paragraph{Pre-training.}
Our Simple I-JEPA and \method models are ViT-B/16~\cite{dosovitskiy2021image} encoders pre-trained on ImageNet-1K~\cite{deng2009imagenet} for 400 epochs on one node with four H100 GPUs, with AdamW~\cite{loshchilov2019decoupled} ($\beta=(0.9,0.95)$), batch size 8192, peak learning rate $4.8\times10^{-3}$ with 40 warm-up epochs and cosine decay, weight decay 0.1, gradient clipping at 3.0 and bfloat16.
The base augmentation is a random resized crop (scale 0.2--1) and a horizontal flip, shared by all views of an image; the corruptions of \cref{sec:denoising} are applied on top of it.
Simple I-JEPA and \method use identical hyper-parameters; the only hyper-parameters we varied are the two specific to \method, the corruption probability $p$ and the weight $\lambda$ of the second-stage loss (\cref{tab:hparam-phijepa}), while the shared ones (learning rate and its schedule, EMA momentum, weight decay, masking and predictor) were not tuned for \method.

\paragraph{Masking and predictor.}
A $224\times224$ image gives $L=196$ patches of $16\times16$ pixels.
The context contains $|\mathcal{C}|=49$ patches (25\% of the image), the square block $\mathcal{B}$ covers $9\times9$ patches, 49 further patches are left out, and the targets are the 81 block patches and 17 scattered patches ($|\mathcal{T}|=98$).
The linear input map of the predictor is implemented as two linear layers without a non-linearity in between. The predictor sequence contains the student's [CLS] token, the 49 context tokens and mask tokens at all 147 non-context positions (including the 49 unused ones, which receive no loss). Its block has two pre-norm attention layers over this whole sequence (the first with separate LayerNorms for the queries and for the keys and values), followed by an MLP of ratio 4, each with a residual connection; the predictor runs in fp32.
The EMA momentum is 0.99, 0.999 and 0.9999 in the first half, the third quarter and the last quarter of training.

\paragraph{Corruptions and second stage.}
Each view enters the corruption pipeline ($p=1$); since each operation below is applied with its own probability, a view can still be left almost unchanged. We chose $p=1$ among the settings of \cref{tab:hparam-phijepa} (selected on the same evaluations that we report; $p=0.5$ gives similar results), and the ablations of \cref{tab:ablation} use $p=0.5$.
The pipeline applies colour jitter (probability 0.8; brightness and contrast factors in $[0.6,1.4]$, saturation in $[0.8,1.2]$), grayscale conversion (0.2), Gaussian blur (0.5; $\sigma\in[0.1,2]$), solarisation (0.2) and additive Gaussian noise (0.2; standard deviation 0.05).
The CA1 head $g$ is a two-layer MLP of widths $768\to2048\to768$, and $\lambda=0.5$.
On one node with four H100 GPUs, 400 epochs take 21.2 hours (85 GPU-hours) for Simple I-JEPA and 25.4 hours (102 GPU-hours) for \method.

\paragraph{Evaluation.}
All encoders are frozen, and we use final-layer patch tokens without the [CLS] token; for ViT-L, the feature block is chosen as described in \cref{sec:suppl-vitl}. For our models we use the EMA encoder; public baselines use their released checkpoints.
\emph{Linear probe:} the mean of the patch tokens (taken before the final LayerNorm and followed by a LayerNorm without learned parameters, as in MAE's global pooling), a BatchNorm and a linear classifier trained with LARS~\cite{you2017scaling} for 90 epochs (MAE protocol~\cite{he2022masked}, but with mean-pooled patch tokens instead of the class token). %
\emph{Attentive probe:} one learned query with a cross-attention block (12 heads; 16 for ViT-L, whose width 1024 is not divisible by 12), an MLP and a linear classifier, trained for 20 epochs with AdamW (learning rate $10^{-3}$, cosine), batch size 1024 and AutoAugment~\cite{cubuk2019autoaugment}, following V-JEPA~\cite{bardes2024revisiting}.
\emph{Linear segmentation:} PASCAL VOC 2012~\cite{everingham2009pascal}; the patch tokens of 448$\times$448 images (28$\times$28 grid) are cached, and a BatchNorm and a linear layer (21 classes) are trained on the 1464 training images for 30 epochs, with the logits upsampled bilinearly; we report val mIoU (1449 images) for the best of three learning rates.
\emph{Baselines.} We compare with methods trained on ImageNet-1K at the same model size, to limit differences in data and model scale. Public checkpoints still differ in training duration, augmentation, and optimisation, so these comparisons do not isolate the objective alone.
We use iBOT~\cite{zhou2021ibot} as a strong ImageNet-1K self-distillation reference and the recent DSeq-JEPA~\cite{he2026dseq} (ECCV 2026), an improved I-JEPA, as the state-of-the-art JEPA with a public ViT-B/16 checkpoint. DINOv2~\cite{oquab2024dinov2} and DINOv3~\cite{simeoni2026dinov3} use substantially larger pre-training datasets and different training pipelines, so we focus on the ImageNet-1K setting.
Public baselines are evaluated with exactly the same protocols. For LeJEPA~\cite{balestriero2025lejepa}, whose official release has no ViT-B, we use a third-party ViT-B/16 reimplementation trained for 300 epochs on ImageNet-1K~\cite{ai2026lejepa}.

\section{Pre-training cost}
\label{sec:suppl-cost}

\begin{table}[t]
\centering
\small
\setlength{\tabcolsep}{3pt}
\begin{adjustbox}{max width=\linewidth}
\begin{tabular}{@{}lcccccc@{}}
\toprule
Method & Arch. & Epochs & Eff.\ epochs & Time & FLOPs & Mem./GPU \\
\midrule
\multicolumn{7}{@{}l}{\emph{full runs, one node with 4$\times$H100}} \\
Simple I-JEPA & ViT-B/16 & 400 & 400 & 21.2\,h (85 GPU-h) & 71 / 36 & 21.0\,GB \\
\method & ViT-B/16 & 400 & 400 & 25.4\,h (102 GPU-h) & 109 / 56 & 23.7\,GB \\
\midrule
\multicolumn{7}{@{}l}{\emph{baseline runs and timing estimates on the same node}} \\
iBOT~\cite{zhou2021ibot}$^\star$ & ViT-B/16 & 400 & 1600 & $\approx$132\,h ($\approx$530 GPU-h), est. & 514 / 263 & 52.3\,GB$^\star$ \\
LeJEPA~\cite{balestriero2025lejepa}$^\circ$ & ViT-B/16 & 400 & -- & $\approx$72\,h ($\approx$290 GPU-h), est. & 316 / 162 & 27.1\,GB$^\circ$ \\
I-JEPA~\cite{assran2023self}$^\diamond$ & ViT-B/16 & 600 & 600 & 56.3\,h (225 GPU-h) & 98 / 75 & 12.2\,GB$^\diamond$ \\
MoCo~v3~\cite{chen2021empirical}$^\dagger$ & ViT-B/16 & 300 & 600 & $\approx$65\,h ($\approx$260 GPU-h), est. & 280 / 108 & -- \\
MAE~\cite{he2022masked}$^\P$ & ViT-B/16 & 1600 & 1600 & $\approx$80\,h ($\approx$320 GPU-h), est. & 56 / 115 & 8.3\,GB$^\P$ \\
\midrule
\multicolumn{7}{@{}l}{\emph{not run on our node; estimated from the authors' reported cost relative to I-JEPA}} \\
DSeq-JEPA~\cite{he2026dseq}$^\S$ & ViT-B/16 & 600 & 600 & $\approx$62\,h ($\approx$248 GPU-h), est. & $\approx$108 / $\approx$83 & $\approx$15\,GB$^\S$, est. \\
\bottomrule
\end{tabular}
\end{adjustbox}
\caption{\textbf{Pre-training cost} (ViT-B/16). Simple I-JEPA, \method and I-JEPA (our pre-training with the official code): measured full runs. Other baselines: per-epoch time measured with the official recipes on the same node and extrapolated to the full schedule, except DSeq-JEPA, which is scaled from its authors' reported overhead relative to I-JEPA; these totals are estimates. FLOPs: training GFLOPs per image / total for the listed schedule ($\times10^{18}$), counted on one training step of each method's code (DSeq-JEPA: I-JEPA scaled by the reported time ratio). Details of each measurement are given below. The evaluated LeJEPA checkpoint was trained for 300 epochs (121$\times10^{18}$ FLOPs, $\approx$216 GPU-h, as in \cref{tab:main}).}
\label{tab:compute}
\end{table}

\begin{figure*}[t]
  \centering
  \includegraphics[width=\linewidth]{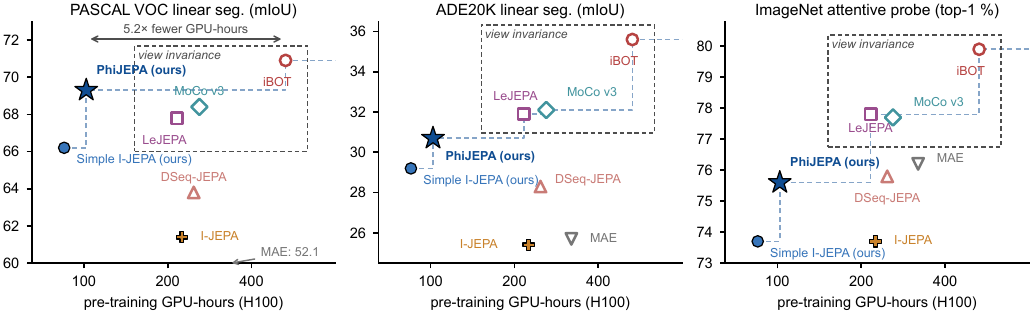}
  \caption{\textbf{Accuracy and pre-training GPU-hours.} The checkpoints of \cref{fig:efficiency} with measured or estimated H100 GPU-hours on our node (\cref{tab:compute}) on the horizontal axes (log scale). Filled markers: measured full runs (I-JEPA: our re-training, \cref{tab:main}). Open markers: estimated costs from official-code timings, a reimplemented step for LeJEPA, or reported overhead relative to I-JEPA for DSeq-JEPA. With GPU-hours, \method needs less than half the time of the evaluated I-JEPA and DSeq-JEPA checkpoints (102 \vs 225 and $\approx$248) and approximately one fifth of iBOT's ($\approx$530).}
  \label{fig:efficiency-gpuh}
\end{figure*}
Relative to Simple I-JEPA, \method adds a teacher forward pass on the independently corrupted view and a token-wise MLP.
This increases training time by 20\% and memory by 13\%, while the student still processes only 49 context patches.
iBOT processes two global and ten local crops per image and is $5.2\times$ slower per epoch on the same hardware, based on first-epoch timing extrapolated to 400 epochs.
The full I-JEPA run takes 56.3 hours on the same four-GPU node, approximately $2.2\times$ the training time of \method (\cref{tab:compute}).

\paragraph{FLOP counts for \cref{tab:compute}.}
We count floating-point operations with PyTorch's \texttt{FlopCounterMode} on one training step (batch of four images, on CPU) of the same code used for the timings: our implementation for Simple I-JEPA and \method, the official code for I-JEPA (with its multi-block masks, averaged over 20 mask draws: 68 context and 130 target tokens per image), iBOT, MAE and MoCo~v3, and the LeJEPA step described below.
We count the forward and backward passes of the trained networks and the forward passes of EMA or momentum teachers; a multiply--accumulate counts as two FLOPs, and element-wise operations, the optimiser and data augmentation are not counted.
Totals multiply the per-image count by the 1.28M ImageNet-1K training images and the number of epochs.
For DSeq-JEPA, we scale the I-JEPA count by the reported training-time ratio ($1.10\times$). This proxy assumes proportionality between runtime and FLOPs, although hardware utilisation and memory traffic can change runtime independently of operation counts. Direct FLOP counting is needed to establish the compute difference from DSeq-JEPA.
Per image, \method costs 109 GFLOPs: 49 visible tokens in the student, two full teacher passes (the corrupted and the clean view) and the predictor and CA1 head; Simple I-JEPA, with one teacher pass, costs 71.
I-JEPA costs a similar 98 GFLOPs per image, since its student encodes more context tokens (68) and its six-block predictor runs for four target blocks.
Measured time per epoch is nonetheless 1.5$\times$ higher for I-JEPA (5.6 \vs 3.8\,min), so GPU-hours and FLOPs give different ratios; we report both.

\paragraph{Timing details for \cref{tab:compute}.}
Effective epochs count all training views under iBOT's multi-crop convention.
For Simple I-JEPA and \method, we measure complete 400-epoch runs with effective batch size 8192 (512 images per GPU on four GPUs, accumulated over four steps) and report peak allocated memory per GPU. $^\star$iBOT uses the official ViT-B/16 code and recipe: 2 global and 10 local crops, fp16 and batch size 128 per GPU.
Its 400-epoch cost is an \emph{estimate} extrapolated from steady-state iterations in the first epoch (19.8\,min per epoch; 0.476\,s per 512 images, compared with 3.8\,min per epoch for \method). $^\circ$LeJEPA has no ImageNet training script in the official release, so we time a reimplementation using its SIGReg loss and invariance term.
The setup uses ViT-B/16, a 512-d embedding, an MLP projector, 2 global views at 224 pixels, 6 local views at 96 pixels, official augmentations, bf16 and batch size 128 per GPU.
Averaged over 350 post-warm-up iterations, each batch of 512 images takes 0.259\,s, or 10.8\,min per epoch; the 400-epoch total is estimated. The evaluated LeJEPA checkpoint uses 300 epochs, for an estimated 216 GPU-hours.

$^\diamond$I-JEPA uses the official code, a ViT-B/16 encoder and the paper's 6-block predictor.
We measure a complete 600-epoch run (our re-training with the DSeq-JEPA recipe, \cref{tab:main}) with 2048 images per step, averaging 5.6\,min per epoch.
Memory is measured in a separate timing run with the official batch size of 128 images per GPU. $^\S$DSeq-JEPA is not timed on our node.
Its authors report 26.5\,h versus I-JEPA's 24.2\,h and 38.2 versus 31.5\,GB peak memory on 8 H200 GPUs with ViT-B/16~\cite{he2026dseq}.
We scale our I-JEPA measurements by these ratios ($1.10\times$ time and $1.21\times$ memory), giving an estimated 6.2\,min per epoch.

$^\dagger$MoCo~v3 uses the official code and README recipe (AdamW, two 224-pixel crops, momentum encoder and AMP), with total batch size 512 rather than 4096.
Including data loading, the first four epochs average 0.31\,s per batch of 512 images, or 13.0\,min per epoch ($3.4\times$ \method).
The 300-epoch total is estimated. $^\P$MAE uses the official code and recipe (mask ratio 0.75, normalised pixel loss and AMP), with batch size 128 per GPU.
The measured average over 19 epochs is 3.0\,min per epoch ($0.79\times$ \method), extrapolated to the 1600-epoch schedule.
DINO's 400-epoch schedule has not been timed.

\section{Hyper-parameters and training dynamics of \method}
\label{sec:suppl-more}
\label{sec:exp-dynamics}

We established the shared training hyper-parameters using Simple I-JEPA as the baseline and transferred them unchanged to \method. The default settings specific to \method ($p$, $\lambda$ and the CA1 head) are fixed for the main-paper comparison.
\Cref{tab:hparam-phijepa} lists the default result and the variants.
We vary the corruption probability $p$ and CA1 loss weight $\lambda$.
All settings shared with Simple I-JEPA---learning rate and schedule, EMA momentum, weight decay, batch size, masking and first predictor---remain unchanged.
All tested variants lie within 68.0--69.9 mIoU on VOC, 30.7--31.2 on ADE20K and 75.1--75.8\% in attentive probing, showing that the gains persist across these tested settings. The single-run results also show differences across tasks.
We choose $p=1$ because it gives similar segmentation and probing to $p=0.5$. %
This choice uses the reported evaluations.

\begin{table}[h]
\centering
\small
\setlength{\tabcolsep}{3pt}
\begin{adjustbox}{max width=\linewidth}
\begin{tabular}{@{}cccccc@{}}
\toprule
$p$ & $\lambda$ & VOC & ADE20K & Lin. & Attn. \\
\midrule
0.5 & 0.5 & 69.2 & 31.2 & 72.4 & 75.8 \\
0.25 & 0.5 & 69.9 & 30.7 & 73.2 & 75.5 \\
\textbf{1} & \textbf{0.5} & 69.3 & 30.7 & 72.0 & 75.6 \\
0.5 & 0.25 & 68.0 & 30.9 & 72.4 & 75.1 \\
0.5 & 1 & 69.1 & 31.1 & 72.8 & 75.5 \\
\bottomrule
\end{tabular}
\end{adjustbox}
\caption{\textbf{Default configuration and variants of \method} (ViT-B/16, ImageNet-1K, 400 epochs). $p$: corruption probability; $\lambda$: weight of the CA1 loss. Bold settings: default \method ($p=1$), used in the main paper. The first row is the configuration with $p=0.5$ used in the ablations; each other row changes one setting of it. VOC and ADE20K: linear segmentation (mIoU); Lin.\ and Attn.: ImageNet linear and attentive probes (top-1).}
\label{tab:hparam-phijepa}
\end{table}
\paragraph{Stability alone does not explain the gain.}
At epoch 50, independent corruption without CA1 already reduces relative feature sensitivity from 0.343 to 0.081, close to 0.077 with CA1.
Yet the second stage improves segmentation at every checkpoint in \cref{tab:training-main}.
From epoch 200 on, the no-CA1 run also falls below Simple I-JEPA, despite its earlier gains.
These results distinguish feature stability from useful semantic content, but do not establish the mechanism linking them.

\begin{table}[t]
\centering
\small
\setlength{\tabcolsep}{4pt}
\begin{tabular}{@{}lccccc@{}}
\toprule
Epoch & 50 & 100 & 200 & 300 & 400 \\
\midrule
Simple I-JEPA & 43.1 & 49.4 & 62.5 & 66.2 & 66.2 \\
Independent, no CA1 & 48.6 & 54.9 & 62.2 & 65.6 & 65.9 \\
\method, $p=0.5$ & \best{51.6} & \best{59.9} & \best{65.0} & \best{69.2} & \best{69.2} \\
\bottomrule
\end{tabular}
\caption{\textbf{The second stage throughout training.} VOC linear segmentation mIoU on frozen full-image EMA features. ViT-B/16, one run per 400-epoch schedule. The corrupted variants use $p=0.5$.}
\label{tab:training-main}
\end{table}

\section{Hidden-region prediction}
\label{sec:suppl-predict}
We measure how much semantic content the predictor recovers in hidden image regions from visible context.
Each VOC validation image uses the Simple I-JEPA mask: 49 visible patches and 98 targets comprising a $9\times9$ block and 17 scattered patches (\cref{sec:simple}).
The predictor estimates the teacher tokens at these target positions.
A linear segmentation head (BatchNorm and a linear layer, 224 pixels) is trained on the teacher's tokens of the clean train images and applied to the predicted tokens; we report the mIoU over the pixels of the target patches only.
The masks are identical for all models, image by image, so the models see the same context and predict the same positions.
For reference, we include I-JEPA ViT-B/16, re-trained with the official code and the DSeq-JEPA recipe (600 epochs; six-block predictor).

\begin{table}[t]
\centering
\small
\setlength{\tabcolsep}{3pt}
\begin{adjustbox}{max width=\linewidth}
\begin{tabular}{@{}lccccc@{}}
\toprule
& MAE$^\dagger$ & I-JEPA$^\S$ & Simple I-JEPA & \method, $p{=}0.5$ & \method \\
\midrule
mIoU on targets: all / far & & & & & \\
\midrule
teacher, full image (reference) & 65.5 / -- & 62.6 / 63.8 & 65.0 / 65.7 & 66.4 / 67.2 & 67.7 / 68.7 \\
\textbf{prediction}, clean input & 20.1 / -- & 32.5 / 33.9 & 44.3 / 43.9 & 48.5 / 47.6 & \best{49.1} / \best{48.4} \\
\textbf{prediction}, corrupted input & 14.9 / -- & -- & 35.8 / 35.4 & \best{44.6} / \best{43.2} & 44.5 / \best{43.2} \\
nearest visible token (interpolation) & 35.3 / -- & 35.8 / 31.6 & 37.9 / 32.6 & 37.7 / 32.4 & 38.1 / 32.9 \\
mean visible token & 8.0 / -- & 10.6 / 9.7 & 9.4 / 8.6 & 9.5 / 8.6 & 11.0 / 10.1 \\
\midrule
prediction $-$ nearest visible, clean & $-$15.3 / -- & $-$3.3 / +2.4 & +6.4 / +11.3 & +10.9 / +15.2 & \best{+11.0} / \best{+15.4} \\
prediction / teacher & 0.31 & 0.52 & 0.68 & \best{0.73} & \best{0.73} \\
cosine to teacher, clean / corrupted & 0.64 / 0.53 & 0.63 / -- & 0.84 / 0.70 & 0.84 / 0.81 & 0.83 / 0.81 \\
\bottomrule
\end{tabular}
\end{adjustbox}
\caption{\textbf{Semantics of the predicted tokens} (VOC val, ViT-B/16). Differences are computed from the unrounded scores. mIoU of a linear segmentation head trained on the teacher's tokens, evaluated on the pixels of the masked target patches; all models use the same context and target patches. Prediction: the predictor output (for \method the CA3 output; the CA1 output gives 48.9 and 44.3, and 48.5 and 44.4 for $p{=}0.5$). \method, $p{=}0.5$: the variant of \cref{tab:hparam-phijepa} in which each view is corrupted with probability 0.5. Corrupted input: one fixed draw of the training corruption applied to the student input; the teacher and the labels are clean. Far: the 61\% of the target patches at least two patches away from every visible patch (mostly the interior of the block). Nearest visible token: the teacher's token of the closest visible patch, copied to each target (an interpolation baseline; it uses the clean teacher tokens also in the corrupted condition, 37.6 / 32.3 for \method and 36.8 / 31.5 for Simple I-JEPA). Mean visible token: the mean of the teacher's tokens at the visible positions, used for every target. Both copying baselines use contextualised tokens from a full-image teacher pass, which can contain information from the hidden regions. $^\S$Re-trained with the official I-JEPA code and the DSeq-JEPA recipe (600 epochs, \cref{tab:main}); six-block predictor; teacher tokens after the parameter-free LayerNorm of its training targets. $^\dagger$Official MAE ViT-B/16 with its pixel decoder: it reconstructs the hidden patches from the same 49 visible patches, and the completed image (visible pixels and reconstruction) is encoded by the MAE encoder; the head is trained on the MAE encoder's tokens of the full images. Examples in \cref{fig:masked-pred}.}
\label{tab:masked-pred}
\end{table}

\Cref{tab:masked-pred} shows that the predictions of \method carry the most semantic content: 49.1 mIoU, against 44.3 for Simple I-JEPA and 32.5 for I-JEPA.
Relative to Simple I-JEPA, \method gains 4.8 mIoU on predicted tokens and 2.7 mIoU on full-image teacher tokens. Its predicted-to-teacher mIoU ratio rises from 0.68 to 0.73.
The larger gain on predicted tokens is consistent with improved semantic completion. Each model uses a head trained in its own teacher feature space, so the score reflects both encoder quality and compatibility between predicted and teacher tokens.

We compare these predictions with nearest-visible-token copying.
For the default \method checkpoint, copying the teacher's token of the nearest visible patch reaches 38.1 mIoU on all targets and 32.9 on far targets, at least two patches away from every visible patch.
\method exceeds this baseline by 11.0 mIoU on all targets and by 15.4 on the far ones, where its accuracy hardly drops (48.4), whereas Simple I-JEPA exceeds its corresponding baseline by 6.4 and 11.3 mIoU. These gains persist in the interior of large hidden regions.
The predictions of \method are also more stable when the visible input is corrupted, dropping by 4.6 mIoU against 8.5 for Simple I-JEPA, consistent with the first stage being trained to predict the content shared by independently corrupted views.

Under identical masks, the I-JEPA predictor recovers about half its teacher's segmentation accuracy (0.52).
It falls below interpolation on all targets ($-$3.3 mIoU) but exceeds it on far targets (+2.4); its own multi-block mask gives a similar score (32.9 mIoU).
MAE reconstructs the hidden pixels from the same visible patches.
Re-encoding the completed image gives 20.1 mIoU, below nearest-visible-token copying (35.3).
This evaluates the entire reconstruction--re-encoding pipeline, including the decoder and its compatibility with the encoder.

Corrupting each view only with probability 0.5 during pre-training gives similar predictions (48.5 mIoU on clean inputs and 44.6 on corrupted ones).
At test time, CA1 and CA3 give similar accuracy (48.9 \vs 49.1).
Together with the ablations, this is consistent with the second stage helping to train the representation and first predictor, without improving this metric at inference.

\begin{figure*}[t]
  \centering
  \includegraphics[width=0.75\linewidth]{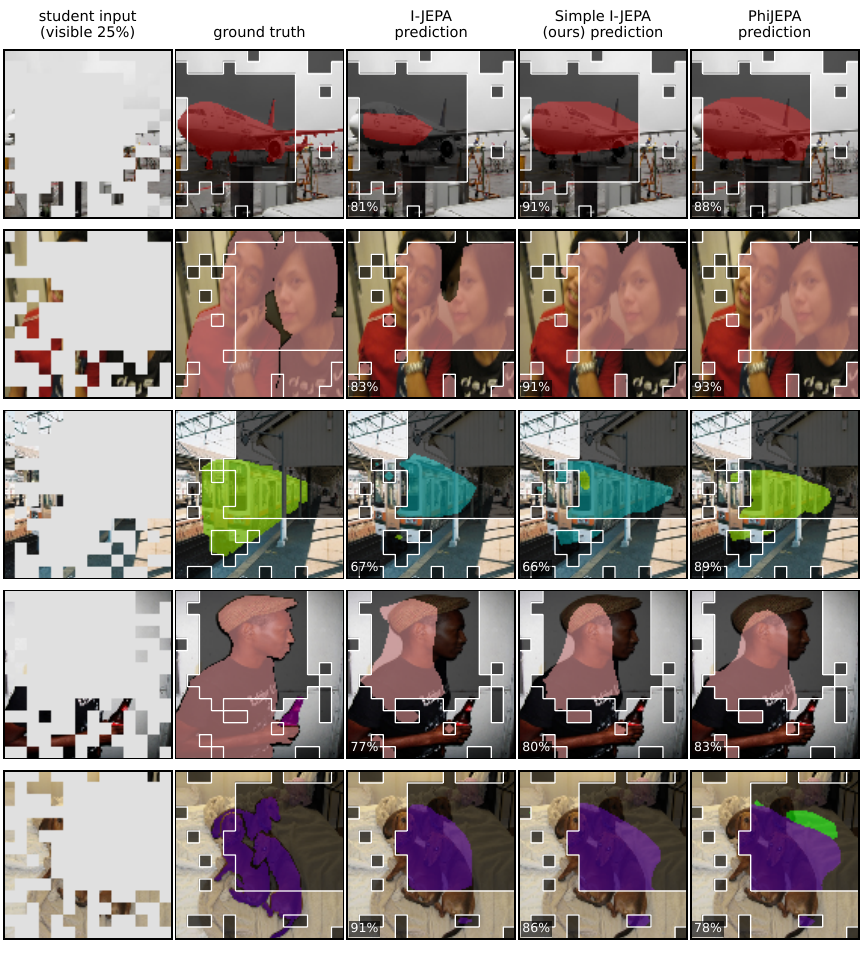}
  \caption{\textbf{Predicting hidden regions.} The student sees only the visible 25\% of the image (first column; hidden patches in grey), and the predictor predicts the tokens of the target patches (outlined: the $9\times9$ block and the scattered targets); the masks are identical for the three models. The predicted tokens are segmented by the linear head trained on each model's own teacher tokens, and the classes are overlaid on the image inside the target patches (darkened); the number gives the pixel accuracy over the target patches. The image content inside the targets is shown for reference only and is not seen by the student. The examples are images among the first 40 VOC val images with a large foreground area in the targets and different classes; the last row is a failure case of \method. I-JEPA: re-trained with the official code and the DSeq-JEPA recipe (600 epochs).}
  \label{fig:masked-pred}
\end{figure*}

\section{Robustness and transfer}
\label{sec:suppl-robust}

\begin{table*}[t]
\centering
\small
\setlength{\tabcolsep}{7pt}
\begin{adjustbox}{max width=\linewidth}
\begin{tabular}{@{}lcccc c cccc@{}}
\toprule
& \multicolumn{4}{c}{linear probe} && \multicolumn{4}{c}{attentive probe} \\
\cmidrule(lr){2-5}\cmidrule(lr){7-10}
& clean & keep & occl. & corr. && clean & keep & occl. & corr. \\
Method & & 25\% & 50\% & sev.\ 3 && & 25\% & 50\% & sev.\ 3 \\
\midrule
DINO~\cite{caron2021emerging} & 71.2 & 44.3 & 26.8 & 58.2 && 78.5 & 58.9 & 55.5 & 68.5 \\
iBOT~\cite{zhou2021ibot} & 73.7 & 55.5 & 37.6 & \best{64.2} && \best{79.7} & 63.0 & 56.0 & \best{71.5} \\
LeJEPA$^\ddagger$~\cite{balestriero2025lejepa} & 70.5 & 51.8 & 22.0 & 57.1 && 78.1 & 59.9 & 52.8 & 66.4 \\
\midrule
I-JEPA$^\S$~\cite{assran2023self} & 71.0 & 62.2 & 47.3 & 43.9 && 73.7 & 65.0 & 54.4 & 51.4 \\
DSeq-JEPA~\cite{he2026dseq} & \best{74.0} & 64.1 & \best{54.4} & 54.5 && 75.8 & 65.2 & \best{58.1} & 59.8 \\
Simple I-JEPA & 72.2 & \best{66.4} & 46.2 & 49.4 && 73.7 & 67.6 & 50.9 & 54.3 \\
\method & 71.9 & 64.8 & 38.1 & 60.9 && 75.7 & \best{67.9} & 47.1 & 64.9 \\
\bottomrule
\end{tabular}
\end{adjustbox}
\caption{\textbf{Frozen-feature robustness under photometric corruption and partial observation.} Frozen encoders with their trained probes; ImageNet top-1 on the same 10k validation images and perturbations. Keep: only a random 25\% of tokens enter the encoder. Occl.: a square covering 50\% of the image is filled with the mean colour. Corr.: average over Gaussian noise, blur, JPEG, contrast and brightness at severity 3. Clean values use this subset and differ from \cref{tab:main}. $^\ddagger$Third-party LeJEPA checkpoint. $^\S$Re-trained with the official I-JEPA code and the same recipe as DSeq-JEPA (as in \cref{tab:main}). Bold: best in each column.}
\label{tab:robust}
\end{table*}

\paragraph{Different responses to appearance and missing evidence.}
\Cref{tab:robust} evaluates the same images and perturbations with both linear and attentive probes.
With attentive pooling, \method improves corruption accuracy over Simple I-JEPA from 54.3\% to 64.9\%, but remains below iBOT's 71.5\%.
Keeping only 25\% of patch tokens leaves masked latent predictors stronger than the view-invariance baselines, by at least 6.7 points with the linear probe but by at least 2.0 points with the attentive probe.
A filled image occluder is a different intervention: it enters the encoder as image content, whereas token dropping explicitly removes positions.
Here the default \method falls below Simple I-JEPA (47.1\% versus 50.9\% attentive accuracy), and DSeq-JEPA has the highest attentive accuracy (58.1\%).
These results preclude a general claim of improved robustness to all missing observations.

\paragraph{Transfer and label-free downstream tasks.}
\Cref{tab:transfer} evaluates the frozen encoders on further tasks, all on CPU with the same features for every model.
\emph{Linear transfer}: we fit logistic regression to standardised mean patch tokens after the final LayerNorm, using a 224 px centre crop as in the ImageNet linear probe.
L-BFGS trains the classifier, with an L2 penalty selected from $\{10^{-7},\dots,10^{-1}\}$ on a validation split, followed by retraining on train+val.
We evaluate DTD~\cite{cimpoi2014describing} split 1, Flowers-102~\cite{nilsback2008automated}, Oxford-IIIT Pets~\cite{parkhi2012cats} (20\% of trainval held out), CIFAR-100~\cite{krizhevsky2024learning} (10\% of train held out, upsampled images), VOC07 classification~\cite{everingham2009pascal} (mAP, ignoring difficult labels) and ImageNet with 1\% labels using the subset of~\cite{chen2020simple}.
\emph{DAVIS-2017}~\cite{tuset20172017} semi-supervised video object segmentation by label propagation with the patch tokens, following DINO's protocol~\cite{caron2021emerging} (480 px, 7 preceding frames, top-5 affinities within a 12-patch neighbourhood), scored with J\&F on the 30 validation videos; our implementation reproduces the published J\&F of DINO ViT-B/16 (62.3).
\emph{Unsupervised object discovery}: we run TokenCut~\cite{wang2022self} on last-layer attention keys and report CorLoc on VOC07 trainval.
TokenCut thresholds key cosine similarity at 0.2.
For several JEPA checkpoints and iBOT, most similarities exceed this threshold, making the graph nearly complete and reducing CorLoc to $\approx$1\%; LOST~\cite{simeoni2021localizing} fails similarly.
We also report results after centring keys per image, although this reduces DINO and MoCo~v3 performance.
Neither choice treats all encoders equally, so both are reported.

\begin{table*}[t]
\centering
\small
\setlength{\tabcolsep}{3pt}
\begin{adjustbox}{max width=\textwidth}
\begin{tabular}{@{}lcccccccc c cc@{}}
\toprule
& \multicolumn{2}{c}{pre-training} & \multicolumn{6}{c}{linear transfer} & DAVIS & \multicolumn{2}{c}{TokenCut CorLoc} \\
\cmidrule(lr){2-3}\cmidrule(lr){4-9}\cmidrule(lr){11-12}
Method & Ep. & GPU-h & DTD & Flowers & Pets & CIFAR-100 & VOC07 (mAP) & IN 1\% & J\&F & original & centered \\
\midrule
\multicolumn{12}{@{}l}{\emph{view invariance (contrastive / self-distillation / multi-view JEPA)}} \\
DINO~\cite{caron2021emerging} & 400 & -- & 75.7 & \best{95.4} & 88.1 & 80.7 & 84.0 & 46.1 & 62.3 & \best{68.6} & 38.9 \\
iBOT~\cite{zhou2021ibot} & 400 & $\approx$530 & 75.8 & 93.3 & 80.8 & \best{83.0} & 86.2 & 47.1 & 62.5 & 1.1 & 33.3 \\
MoCo~v3~\cite{chen2021empirical} & 300 & $\approx$260 & 76.5 & 95.0 & 86.1 & 82.6 & 84.3 & 48.9 & 61.4 & 67.1 & 43.4 \\
LeJEPA$^\ddagger$~\cite{balestriero2025lejepa} & 300 & $\approx$216 & \best{78.2} & 95.2 & 83.7 & 78.7 & 85.4 & 47.3 & \best{63.6} & 26.5 & 38.5 \\
\midrule
\multicolumn{12}{@{}l}{\emph{masked prediction: pixels (MAE) and latents (JEPAs)}} \\
MAE~\cite{he2022masked} & 1600 & $\approx$320 & 71.7 & 88.9 & 80.8 & 69.2 & 81.6 & 36.9 & 51.1 & 45.5 & 34.8 \\
I-JEPA$^\S$~\cite{assran2023self} & 600 & 225 & 66.4 & 87.8 & 89.2 & 76.5 & 83.5 & 56.1 & 55.8 & 41.0 & 46.0 \\
DSeq-JEPA~\cite{he2026dseq} & 600 & $\approx$248 & 70.0 & 90.8 & \best{91.1} & 80.1 & \best{87.2} & \best{61.0} & 57.5 & 50.9 & \best{50.4} \\
Simple I-JEPA & 400 & 85 & 66.8 & 86.7 & 87.5 & 76.5 & 86.3 & 57.1 & 53.2 & 1.1 & 49.0 \\
\method & 400 & 102 & 71.3 & 91.6 & 86.9 & 78.4 & 86.3 & 54.7 & 58.7 & 29.9 & 49.8 \\
\midrule
\multicolumn{12}{@{}l}{\emph{larger model, for reference (ViT-H/14, 632M parameters, $7\times$ ViT-B/16)}} \\
I-JEPA~\cite{assran2023self} \textbf{ViT-H/14} & 300 & -- & 71.4 & 90.3 & 90.3 & 84.5 & 88.8 & 66.3 & 59.6 & 0.4 & 28.8 \\
\bottomrule
\end{tabular}
\end{adjustbox}
\caption{\textbf{Transfer and label-free downstream tasks} with frozen encoders (ViT-B/16 unless stated). Linear transfer: top-1 (\%) except VOC07 (mAP). IN 1\%: 13 labelled images per class, logistic regression on the mean patch token (lower than published semi-supervised results, which fine-tune the encoder). DAVIS-2017: J\&F by label propagation. TokenCut: CorLoc on VOC07 trainval with the original keys and with keys centered per image. Ep.: pre-training epochs of the evaluated checkpoint; GPU-h: pre-training H100 GPU-hours on one node with four GPUs, measured for Simple I-JEPA, \method and I-JEPA and estimated ($\approx$) from measured timings of official code (iBOT, MoCo~v3, MAE), a reimplemented training step (LeJEPA) or from the authors' reported cost (DSeq-JEPA), \cf \cref{tab:compute}; --: not measured. $^\ddagger$Third-party reimplementation~\cite{ai2026lejepa}. $^\S$Re-trained with the official I-JEPA code and the same recipe as DSeq-JEPA (as in \cref{tab:main}). Best ViT-B/16 model per column in bold; I-JEPA ViT-H/14, a $7\times$ larger model, is shown separately for reference and is not included in the comparison.}
\label{tab:transfer}
\end{table*}

\section{Completing occluded PushT observations}
\label{sec:suppl-occlusion}
A latent world model can predict how frozen visual features evolve under actions~\cite{zhou2025dino}.
Occlusion changes these features and can remove information needed for planning.
A JEPA's pre-trained predictor provides a way to estimate hidden tokens from visible patches. We evaluate the agreement of these estimates with unoccluded features.
We test this on PushT~\cite{chi2023diffusion}, the manipulation task of DINO-WM, without any training on PushT.
We take 300 random frames of the DINO-WM validation trajectories (224 pixels, $14\times14$ patch grid) and occlude a square of cells, either centred on the T-shaped block that has to be pushed ($4\times4$ cells) or at a random position ($4\times4$ or $6\times6$ cells).
For each frozen encoder, the tokens of the occluded cells are then obtained in five ways: from the occluded image as it is; by copying the token of the nearest visible cell; by inpainting the pixels, with the classical method of Telea~\cite{telea2004image} or with the official MAE ViT-B/16~\cite{he2022masked}, and encoding the result; or, for the JEPAs, with the encoder's own predictor from the visible cells.
The tokens of the visible cells are those of the occluded image in all cases, and we compare the tokens of the occluded cells with those of the unoccluded frame.

\begin{table}[t]
\centering
\small
\setlength{\tabcolsep}{3pt}
\begin{adjustbox}{max width=\linewidth}
\begin{tabular}{@{}llccc@{}}
\toprule
encoder & completion & on block $4\times4$ & random $4\times4$ & random $6\times6$ \\
\midrule
\multirow{5}{*}{\shortstack[l]{\method,\\ $p{=}0.5$}} & none (occluded image) & 0.796 & 0.806 & 0.822 \\
 & nearest visible token & 0.857 & 0.923 & 0.875 \\
 & pixels, Telea & 0.814 & 0.930 & 0.886 \\
 & pixels, MAE & 0.790 & 0.899 & 0.862 \\
 & \textbf{own predictor} & \best{0.900} & \best{0.972} & \best{0.948} \\
\midrule
\multirow{5}{*}{\method} & none (occluded image) & 0.767 & 0.860 & 0.855 \\
 & nearest visible token & 0.825 & 0.923 & 0.877 \\
 & pixels, Telea & 0.775 & 0.914 & 0.876 \\
 & pixels, MAE & 0.713 & 0.860 & 0.792 \\
 & \textbf{own predictor} & \best{0.897} & \best{0.972} & \best{0.950} \\
\midrule
\multirow{5}{*}{Simple I-JEPA} & none (occluded image) & 0.686 & 0.789 & 0.792 \\
 & nearest visible token & 0.819 & 0.927 & 0.878 \\
 & pixels, Telea & 0.815 & 0.946 & 0.888 \\
 & pixels, MAE & 0.773 & 0.858 & 0.795 \\
 & \textbf{own predictor} & \best{0.888} & \best{0.968} & \best{0.940} \\
\midrule
\multirow{5}{*}{I-JEPA$^\S$} & none (occluded image) & 0.396 & 0.419 & 0.486 \\
 & nearest visible token & 0.575 & 0.722 & 0.676 \\
 & pixels, Telea & 0.574 & 0.848 & 0.757 \\
 & pixels, MAE & 0.534 & 0.589 & 0.550 \\
 & \textbf{own predictor} & \best{0.695} & \best{0.859} & \best{0.813} \\
\midrule
\multirow{4}{*}{DINOv2 ViT-B/14} & none (occluded image) & 0.469 & 0.359 & 0.360 \\
 & nearest visible token & \best{0.508} & \best{0.534} & \best{0.499} \\
 & pixels, Telea & 0.456 & 0.487 & 0.483 \\
 & pixels, MAE & 0.452 & 0.363 & 0.356 \\
\bottomrule
\end{tabular}
\end{adjustbox}
\caption{\textbf{Completing occluded PushT observations} (300 validation frames of DINO-WM, no training on PushT). Mean cosine similarity between the completed tokens of the occluded cells and the tokens of the unoccluded frame, for each frozen encoder; best completion of each encoder in bold. Occluders: $4\times4$ cells centred on the T block, or $4\times4$ / $6\times6$ cells at a random position (8\% / 18\% of the image). For \method, the predictor output is that of CA3; the CA1 head gives 0.897, 0.972 and 0.949 (0.900, 0.971 and 0.947 for $p{=}0.5$, each view corrupted with probability 0.5 during pre-training). Cosine similarity is evaluated within each encoder's feature space. Cross-encoder differences also depend on feature geometry. DINOv2 has no predictor. $^\S$Re-trained with the official I-JEPA code and the DSeq-JEPA recipe (600 epochs, \cref{tab:main}).}
\label{tab:occlusion}
\end{table}

\Cref{tab:occlusion} shows that, for all evaluated JEPA encoders, the encoder's own predictor completes the occluded cells better than every alternative, including completion with the pre-trained MAE, although the predictors were trained on ImageNet only.
The margin is large when the occluder hides the block itself, the object whose pose the task depends on: the predictor of \method reaches a cosine similarity of 0.897 with the unoccluded tokens, against 0.825 for the best alternative, while the tested pixel-inpainting methods give lower feature agreement.
The pre-trained predictors improve feature agreement over the tested pixel-completion methods without additional training on PushT, at the cost of an extra predictor pass.
The $p=0.5$ variant behaves similarly (0.900, 0.972 and 0.948 with its predictor).
These comparisons establish agreement within each encoder's feature space.
\subsection{State read-out from occluded observations}
\label{sec:exp-occlusion}
We separately evaluate whether the encoder and completion method preserve physical state, without training a dynamics model.
We sample 300 frames from the DINO-WM validation trajectories of PushT~\cite{chi2023diffusion,zhou2025dino}, without additional training on PushT.
A $4\times4$-patch occluder covers the T-shaped block on the $14\times14$ grid.
For JEPA completion, the predictor estimates the hidden tokens from visible patches.
Visible tokens are taken from the occluded image in every completion condition.
The occlusion mask is supplied to the predictor.

For each completed observation, we retrieve the nearest of 3,300 unoccluded frames using the squared latent distance used by DINO-WM's planner.
A match satisfies the reported PushT state criterion when the joint position error of the agent and the block is below 20 units in PushT environment coordinates and the block angle is within $20^\circ$.
The resulting rate measures the state information preserved by the representation and completion pipeline.

\begin{table}[t]
\centering
\small
\setlength{\tabcolsep}{5pt}
\begin{tabular}{@{}llr@{}}
\toprule
Encoder & Completion & State match (\%) \\
\midrule
\method & None & 16 \\
\method & Pixel inpainting, Telea & 33 \\
\method & Pixel completion, MAE & 29 \\
\method & Nearest visible token & 39 \\
\method & Own predictor & \best{53} \\
\midrule
Simple I-JEPA & Best tested completion & 16 \\
DINOv2-B & None & 42 \\
DINOv2-B & Best tested completion & 44 \\
\bottomrule
\end{tabular}
\caption{\textbf{State retrieval with the task object occluded.}
Fraction of the 300 queries whose retrieved clean frame satisfies the position-and-angle criterion.
\method uses the default $p=1$ checkpoint; its own predictor is the first predictor followed by the CA1 head, as in planning.
These are frame-retrieval results.
Feature-agreement comparisons for all occlusion settings and completion methods are in \cref{tab:occlusion}.}
\label{tab:state-retrieval}
\end{table}

\begin{figure*}[t]
  \centering
  \includegraphics[width=0.88\linewidth,trim=0 223 0 0,clip]{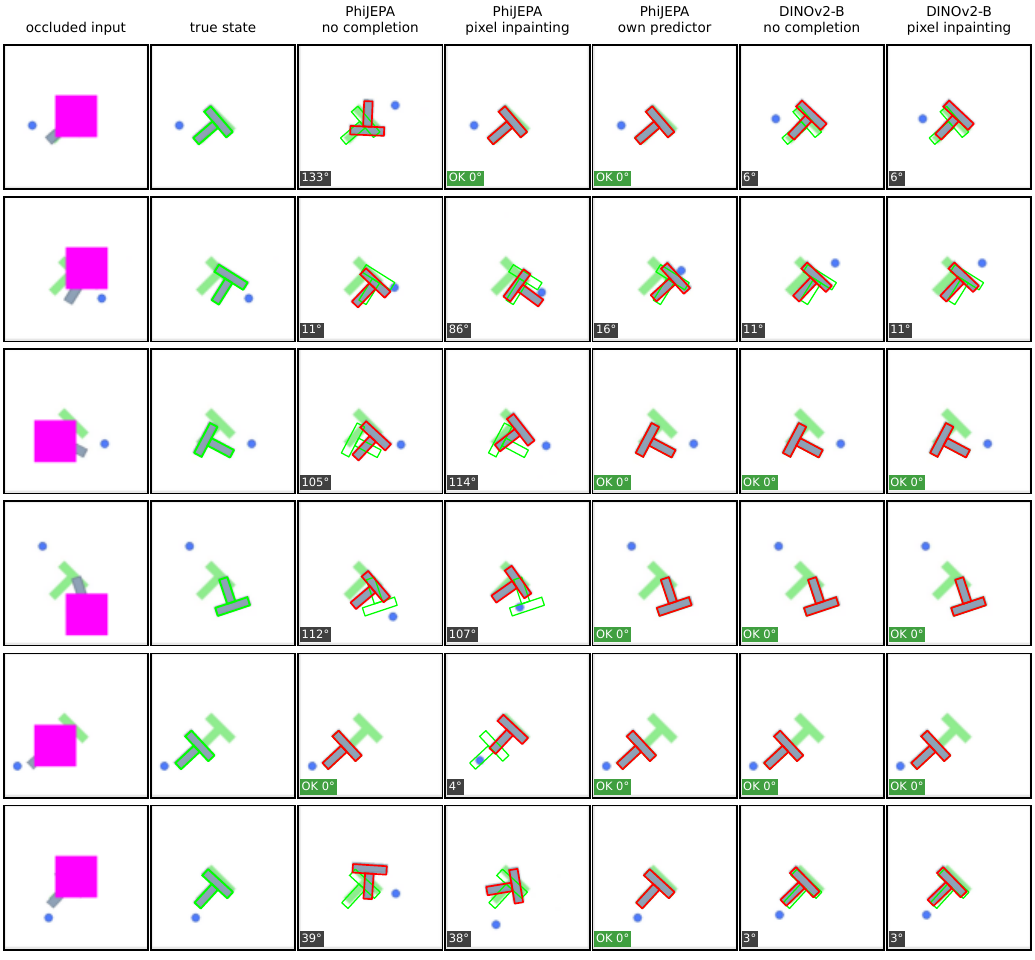}\\[2pt]
  \includegraphics[width=0.88\linewidth,trim=0 0 0 384,clip]{fig/pusht_occlusion.pdf}
  \caption{\textbf{State retrieval from occluded PushT images.} Rows 1--3 are the first three random queries. Row 4 is query 6, selected to illustrate a state match with the own predictor while all other displayed alternatives fail. Results are retrieved clean frames, not generated images or rollouts. The T block is covered by a $4\times4$-patch occluder. Red/green: retrieved/true block pose. Corners give angle error; ``OK'' denotes a match under the joint position-and-angle criterion. The light green T is the goal pose. \method uses $p=1$; there is no PushT training or action-conditioned prediction.}
\label{fig:pusht-occlusion}
\end{figure*}

\method's predictor raises the state-match rate from 16\% to 53\%, exceeding the tested alternatives for the same encoder (\cref{tab:state-retrieval}).
The best tested DINOv2-B completion reaches 44\%.
Examples are shown in \cref{fig:pusht-occlusion}.
The feature-agreement evaluation above also includes random occluders.
Those cosine scores compare completion methods within each encoder's feature space; feature geometry can differ across encoders.
The state criterion provides an additional evaluation in shared physical coordinates.

In \cref{fig:pusht-occlusion}, the occluder often changes the retrieved block orientation when the tokens are left uncompleted.
Pixel inpainting does not consistently recover the missing block, whereas \method's own predictor yields a state match in three of the four displayed queries.
These examples illustrate the aggregate result; they do not establish generalisation to unseen trajectories.

\begin{figure*}[t]
  \centering
  \includegraphics[width=0.95\linewidth]{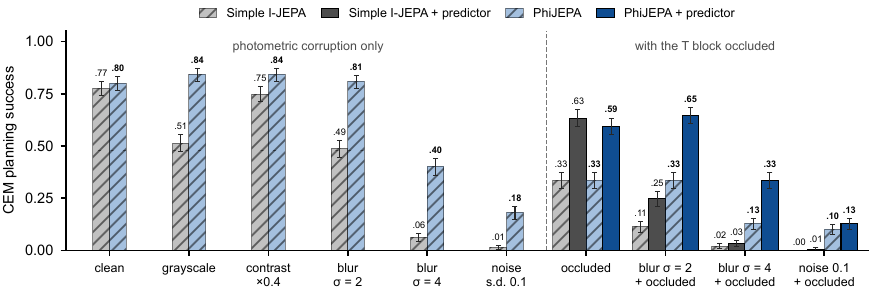}
  \caption{\textbf{CEM planning with \method and Simple I-JEPA} (PushT). Hatched: no completion; solid: completion of the occluded tokens with each encoder's own predictor (only where the T block is occluded). DINO-WM world models after 15 epochs with the same training schedule and batch size; mean over three planning seeds of 50 episodes each; error bars: binomial s.e. Values for the conditions shown in \cref{tab:corruption} match the table.}
  \label{fig:phi-vs-simple}
\end{figure*}

\paragraph{Planning with \method and Simple I-JEPA.}
\Cref{fig:phi-vs-simple} compares \method with Simple I-JEPA, which shares the masked-prediction pipeline but uses neither photometric corruption nor the second prediction stage, using the CEM planning protocol of \cref{tab:corruption}, including the grayscale, contrast and noise conditions not shown there. %
With complete observations the two encoders plan similarly (0.80 and 0.77), and under occlusion alone both predictors restore planning (0.59 and 0.63).
Under photometric corruption without occlusion, \method retains more success in every tested condition (grayscale 0.84 \vs 0.51, blur $\sigma=2$ 0.81 \vs 0.49, blur $\sigma=4$ 0.40 \vs 0.06).
When the occluded observation is also corrupted, \method's predictor still raises success (0.65 with blur $\sigma=2$, 0.33 with $\sigma=4$), whereas Simple I-JEPA's predictor gives at most 0.25.
Additive noise with standard deviation 0.1 severely limits planning for both encoders (at most 0.18).

%% file: sec/X_vitl.tex
\section{Scaling to ViT-L (preliminary)}
\label{sec:suppl-vitl}
The results in this section are preliminary: \method was pre-trained once with ViT-L/16, using a recipe that was not tuned further, and baselines at this scale are left for a future version of this paper.
We used the 400-epoch schedule with effective batch size 8192, peak learning rate $2.4\times10^{-3}$, gradient clipping 1.0 and initial EMA momentum 0.996. These settings were chosen for ViT-L and differ from the ViT-B recipe of \cref{sec:setup}.

The ViT-L encoder is strongest a few blocks before the last one.
We therefore choose the feature block with a rule that does not use the segmentation data: the block with the highest ImageNet $k$-NN accuracy (20-NN on the patch-mean features of 10 training images per class, 5{,}000 validation images).
This rule selects block 22 of 24 (46.9\% $k$-NN accuracy, \vs 37.5\% for the last block, both at epoch 399).
The last block has a low effective rank (102, \vs 321 for block 22), which is consistent with its lower scores (66.0 VOC mIoU and 76.3\% attentive-probe accuracy, both at epoch 399).

With block 22, \method reaches 73.1 VOC mIoU and 33.3 ADE20K mIoU, compared with 69.3 and 30.7 for ViT-B (\cref{tab:vitl}), and 78.7\% attentive-probe accuracy, compared with 75.6\%.
The same selection rule picks block 11 of 12 for ViT-B, which gives 69.2 VOC mIoU; the VOC gain over these block-11 features (3.9 mIoU) is therefore not explained by the block selection alone.
These results suggest that \method benefits from a larger encoder. Because they come from a single run with a different recipe, further evaluation is needed to establish the scaling trend.
For reference, the public I-JEPA ViT-H/14 checkpoint (632M parameters, about twice ViT-L/16, pre-trained for 300 epochs) reaches 68.5 VOC mIoU, 30.9 ADE20K mIoU and 80.3\% attentive-probe accuracy from its last block under the same protocols. \method with ViT-L is thus higher on both segmentation benchmarks and lower in attentive-probe classification, with 0.61$\times$ the training FLOPs (182 \vs 298$\times10^{18}$).

\begin{table}[h]
\centering
\small
\setlength{\tabcolsep}{4pt}
\begin{tabular}{@{}lcccccc@{}}
\toprule
Model & Ep. & FLOPs ($10^{18}$) & Feature block & VOC mIoU & ADE mIoU & Attn.\ (\%) \\
\midrule
\method ViT-B/16 & 400 & 56 & last (12) & 69.3 & 30.7 & 75.6 \\
\method ViT-L/16 & 400 & 182 & 22 & \best{73.1} & \best{33.3} & 78.7 \\
\midrule
I-JEPA ViT-H/14 (public, reference) & 300 & 298 & last (32) & 68.5 & 30.9 & \best{80.3} \\
\bottomrule
\end{tabular}
\caption{\textbf{\method with ViT-B/16 and ViT-L/16, and I-JEPA ViT-H/14 for reference.} \method ViT-B: last block, as in \cref{tab:main}; \method ViT-L: block 22, the block with the highest ImageNet $k$-NN accuracy; I-JEPA ViT-H: last block of the public target encoder. VOC/ADE: linear segmentation at 448 pixels as in \cref{tab:main}; Attn.: ImageNet attentive probe (top-1). FLOPs: total pre-training FLOPs, counted as in \cref{tab:compute} (I-JEPA ViT-H with its official 300-epoch configuration). Bold: best in each column. One pre-training run per \method model.}
\label{tab:vitl}
\end{table}

%% file: main.bib
@article{nguyen2026jepa,
  title = {The {JEPA} Predictor: A Transferable Operator for Occluded Feature Completion},
  author = {William Nguyen and Christopher Nguyen},
  journal = {arXiv preprint arXiv:2607.16274},
  year = {2026},
  litlas_id = {P6AJ7G4T0QBS},
  litlas_boards = {PhiJEPA},
}

@inproceedings{bar2024stochastic,
  title = {Stochastic positional embeddings improve masked image modeling},
  author = {Amir Bar and Florian Bordes and Assaf Shocher and Mido Assran and Pascal Vincent and Nicolas Ballas and Trevor Darrell and Amir Globerson and Yann LeCun},
  booktitle = {International Conference on Machine Learning (ICML)},
  year = {2024},
  litlas_id = {browser\_import:stochastic-positional-embeddings-improve-masked-image-model-19509ca4ba1c8cb2},
  litlas_boards = {PhiJEPA},
}

@inproceedings{hafner2019learning,
  title = {Learning Latent Dynamics for Planning from Pixels},
  author = {Danijar Hafner and Timothy P. Lillicrap and Ian Fischer and Ruben Villegas and David Ha and Honglak Lee and James Davidson},
  booktitle = {International Conference on Machine Learning (ICML)},
  year = {2019},
  litlas_id = {P5FMBQGHQC5Q},
  litlas_boards = {PhiJEPA},
}

@article{kaelbling1998planning,
  title = {Planning and Acting in Partially Observable Stochastic Domains},
  author = {Leslie Pack Kaelbling and Michael L. Littman and Anthony R. Cassandra},
  journal = {Artificial Intelligence},
  volume = {101},
  number = {1-2},
  pages = {99--134},
  year = {1998},
  litlas_id = {PHWHRYZXSVDV},
  litlas_boards = {PhiJEPA},
}

@inproceedings{chi2023diffusion,
  title = {Diffusion Policy: Visuomotor Policy Learning via Action Diffusion},
  author = {Cheng Chi and Siyuan Feng and Yilun Du and Zhenjia Xu and Eric Cousineau and Benjamin Burchfiel and Shuran Song},
  booktitle = {Robotics: Science and Systems (RSS)},
  year = {2023},
  litlas_id = {P5C878JPK21P},
  litlas_boards = {PhiJEPA},
}

@article{telea2004image,
  title = {An Image Inpainting Technique Based on the Fast Marching Method},
  author = {Alexandru C. Telea},
  journal = {Journal of Graphics Tools},
  volume = {9},
  number = {1},
  pages = {23--34},
  year = {2004},
  litlas_id = {P9E9B6T4Q586},
  litlas_boards = {PhiJEPA},
}

@article{hafner2025mastering,
  title = {Mastering diverse control tasks through world models},
  author = {Danijar Hafner and Jurgis Pasukonis and Jimmy Ba and Timothy P. Lillicrap},
  journal = {Nature},
  volume = {640},
  number = {8059},
  pages = {647--653},
  year = {2025},
  litlas_id = {PQ6AHCDA1YS0},
  litlas_boards = {PhiJEPA},
}

@inproceedings{ha2018recurrent,
  title = {Recurrent World Models Facilitate Policy Evolution},
  author = {David Ha and Jürgen Schmidhuber},
  booktitle = {Advances in Neural Information Processing Systems (NeurIPS)},
  year = {2018},
  litlas_id = {PTDEGP83ZCVF},
  litlas_boards = {PhiJEPA},
}

@inproceedings{wang2022self,
  title = {{Self-Supervised} Transformers for Unsupervised Object Discovery using Normalized Cut},
  author = {Yangtao Wang and Xi Shen and Shell Xu Hu and Yuan Yuan and James L. Crowley and Dominique Vaufreydaz},
  booktitle = {IEEE/CVF Conference on Computer Vision and Pattern Recognition (CVPR)},
  year = {2022},
  litlas_id = {P2AV3GST8362},
  litlas_boards = {PhiJEPA},
}

@article{tuset20172017,
  title = {The 2017 {DAVIS} Challenge on Video Object Segmentation},
  author = {Jordi Pont-Tuset and Federico Perazzi and Sergi Caelles and Pablo Arbeláez and Alexander Sorkine-Hornung and Luc Van Gool},
  journal = {arXiv preprint arXiv:1704.00675},
  year = {2017},
  litlas_id = {PRZ6236SW6PB},
  litlas_boards = {PhiJEPA},
}

@inproceedings{parkhi2012cats,
  title = {Cats and dogs},
  author = {Omkar M. Parkhi and Andrea Vedaldi and Andrew Zisserman and C. V. Jawahar},
  booktitle = {IEEE Conference on Computer Vision and Pattern Recognition (CVPR)},
  year = {2012},
  litlas_id = {P5G481799710},
  litlas_boards = {PhiJEPA},
}

@inproceedings{nilsback2008automated,
  title = {Automated Flower Classification over a Large Number of Classes},
  author = {Maria-Elena Nilsback and Andrew Zisserman},
  booktitle = {Indian Conference on Computer Vision, Graphics and Image Processing (ICVGIP)},
  year = {2008},
  litlas_id = {PBR89YVFGQJQ},
  litlas_boards = {PhiJEPA},
}

@inproceedings{cimpoi2014describing,
  title = {Describing Textures in the Wild},
  author = {Mircea Cimpoi and Subhransu Maji and Iasonas Kokkinos and Sammy Mohamed and Andrea Vedaldi},
  booktitle = {IEEE/CVF Conference on Computer Vision and Pattern Recognition (CVPR)},
  year = {2014},
  litlas_id = {P7NGR3ZF44GC},
  litlas_boards = {PhiJEPA},
}

@misc{ai2026lejepa,
  title = {{LeJEPA} {ViT-B/16} pre-trained on {ImageNet-1K} (``{ImageNet-1K} {Self-Supervised} {ViT} Baselines'')},
  author = {{OpenKnowledge AI}},
  howpublished = {Hugging Face model card},
  url = {https://huggingface.co/OK-AI/lejepa-vitb16-pretrain-in1k},
  year = {2026},
  litlas_id = {browser\_import:lejepa-vit-b-16-pre-trained-on-imagenet-1k-imagenet-1k-self-7be617467bc35a38},
  litlas_boards = {PhiJEPA},
}

@article{qiu2025improving,
  title = {Improving Joint Embedding Predictive Architecture with Diffusion Noise},
  author = {Yuping Qiu and Rui Zhu and Ying-cong Chen},
  journal = {arXiv preprint arXiv:2507.15216},
  year = {2025},
  litlas_id = {PCA6Z5HMV8DC},
  litlas_boards = {PhiJEPA},
}

@inproceedings{cubuk2019autoaugment,
  title = {{AutoAugment}: Learning Augmentation Strategies From Data},
  author = {Ekin D. Cubuk and Barret Zoph and Dandelion Mané and Vijay Vasudevan and Quoc V. Le},
  booktitle = {IEEE/CVF Conference on Computer Vision and Pattern Recognition (CVPR)},
  year = {2019},
  litlas_id = {PMFBTJF3DRK2},
  litlas_boards = {PhiJEPA},
}

@inproceedings{vincent2008extracting,
  title = {Extracting and composing robust features with denoising autoencoders},
  author = {Pascal Vincent and Hugo Larochelle and Yoshua Bengio and Pierre-Antoine Manzagol},
  booktitle = {International Conference on Machine Learning (ICML)},
  year = {2008},
  litlas_id = {PK7AQK0SQ3SQ},
  litlas_boards = {PhiJEPA},
}

@inproceedings{chen2025denoising,
  title = {Denoising with a {Joint-Embedding} Predictive Architecture},
  author = {Dengsheng Chen and Jie Hu and Xiaoming Wei and Enhua Wu},
  booktitle = {International Conference on Learning Representations (ICLR)},
  year = {2025},
  litlas_id = {PYF0JHNTQ921},
  litlas_boards = {PhiJEPA},
}

@inproceedings{simeoni2021localizing,
  title = {Localizing Objects with {Self-Supervised} Transformers and no Labels},
  author = {Oriane Siméoni and Gilles Puy and Huy V. Vo and Simon Roburin and Spyros Gidaris and Andrei Bursuc and Patrick Pérez and Renaud Marlet and Jean Ponce},
  booktitle = {British Machine Vision Conference (BMVC)},
  year = {2021},
  litlas_id = {PS5N95ZKAXZM},
  litlas_boards = {PhiJEPA},
}

@article{everingham2009pascal,
  title = {The {Pascal} Visual Object Classes ({VOC}) Challenge},
  author = {Mark Everingham and Luc Van Gool and Christopher K. I. Williams and John M. Winn and Andrew Zisserman},
  journal = {International Journal of Computer Vision},
  volume = {88},
  number = {2},
  pages = {303--338},
  year = {2010},
  litlas_id = {PSEKWADFSRTY},
  litlas_boards = {PhiJEPA},
}

@article{you2017scaling,
  title = {Large Batch Training of Convolutional Networks},
  author = {Yang You and Igor Gitman and Boris Ginsburg},
  journal = {arXiv preprint arXiv:1708.03888},
  year = {2017},
  litlas_id = {PG40E96XH7F0},
  litlas_boards = {PhiJEPA},
}

@inproceedings{loshchilov2019decoupled,
  title = {Decoupled Weight Decay Regularization},
  author = {Ilya Loshchilov and Frank Hutter},
  booktitle = {International Conference on Learning Representations (ICLR)},
  year = {2019},
  litlas_id = {browser\_import:decoupled-weight-decay-regularization-ec198a6c69fd2fa4},
  litlas_boards = {PhiJEPA},
}

@inproceedings{xie2022simmim,
  title = {{SimMIM}: A Simple Framework for Masked Image Modeling},
  author = {Zhenda Xie and Zheng Zhang and Yue Cao and Yutong Lin and Jianmin Bao and Zhuliang Yao and Qi Dai and Han Hu},
  booktitle = {IEEE/CVF Conference on Computer Vision and Pattern Recognition (CVPR)},
  year = {2022},
  litlas_id = {PJY1TSK3FHYF},
  litlas_boards = {PhiJEPA},
}

@inproceedings{zhou2025dino,
  title = {{DINO-WM}: World Models on Pre-trained Visual Features enable Zero-shot Planning},
  author = {Gaoyue Zhou and Hengkai Pan and Yann LeCun and Lerrel Pinto},
  booktitle = {International Conference on Machine Learning (ICML)},
  year = {2025},
  litlas_id = {P3PFCAHHQNH9},
  litlas_boards = {World model; Download; PhiJEPA},
}

@inproceedings{grill2020bootstrap,
  title = {Bootstrap Your Own Latent - A New Approach to {Self-Supervised} Learning},
  author = {Jean-Bastien Grill and Florian Strub and Florent Altché and Corentin Tallec and Pierre H. Richemond and Elena Buchatskaya and Carl Doersch and Bernardo Ávila Pires and Zhaohan Daniel Guo and Mohammad Gheshlaghi Azar and Bilal Piot and Koray Kavukcuoglu and Rémi Munos and Michal Valko},
  booktitle = {Advances in Neural Information Processing Systems (NeurIPS)},
  year = {2020},
  litlas_id = {PSJVM4BZ6SAY},
  litlas_boards = {JEPA; PhiJEPA},
}

@article{garrido2024learning,
  title = {Learning and Leveraging World Models in Visual Representation Learning},
  author = {Quentin Garrido and Mahmoud Assran and Nicolas Ballas and Adrien Bardes and Laurent Najman and Yann LeCun},
  journal = {arXiv preprint arXiv:2403.00504},
  year = {2024},
  litlas_id = {PHPDEZPPR6SA},
  litlas_boards = {JEPA; PhiJEPA},
}

@inproceedings{bao2022beit,
  title = {{BEiT}: {BERT} {Pre-Training} of Image Transformers},
  author = {Hangbo Bao and Li Dong and Songhao Piao and Furu Wei},
  booktitle = {International Conference on Learning Representations (ICLR)},
  year = {2022},
  litlas_id = {PMJDW56D3VM7},
  litlas_boards = {Computer Vision; PhiJEPA},
}

@inproceedings{chen2021empirical,
  title = {An Empirical Study of Training {Self-Supervised} Vision Transformers},
  author = {Xinlei Chen and Saining Xie and Kaiming He},
  booktitle = {IEEE/CVF International Conference on Computer Vision (ICCV)},
  year = {2021},
  litlas_id = {PFVP4YTQ5GG9},
  litlas_boards = {Computer Vision; PhiJEPA},
}

@inproceedings{mo2024connecting,
  title = {Connecting {Joint-Embedding} Predictive Architecture with Contrastive Self-supervised Learning},
  author = {Shentong Mo and Shengbang Tong},
  booktitle = {Advances in Neural Information Processing Systems (NeurIPS)},
  year = {2024},
  litlas_id = {PH8282QV67AE},
  litlas_boards = {JEPA; PhiJEPA},
}

@inproceedings{baevski2023efficient,
  title = {Efficient Self-supervised Learning with Contextualized Target Representations for Vision, Speech and Language},
  author = {Alexei Baevski and Arun Babu and Wei-Ning Hsu and Michael Auli},
  booktitle = {International Conference on Machine Learning (ICML)},
  year = {2023},
  litlas_id = {PE8T5KKNAHGA},
  litlas_boards = {Foundation Model; PhiJEPA},
}

@inproceedings{baevski2022data2vec,
  title = {data2vec: A General Framework for Self-supervised Learning in Speech, Vision and Language},
  author = {Alexei Baevski and Wei-Ning Hsu and Qiantong Xu and Arun Babu and Jiatao Gu and Michael Auli},
  booktitle = {International Conference on Machine Learning (ICML)},
  year = {2022},
  litlas_id = {PTTVKG618Y8M},
  litlas_boards = {Check; PhiJEPA},
}

@inproceedings{zhou2021ibot,
  title = {{iBOT}: Image {BERT} {Pre-Training} with Online Tokenizer},
  author = {Jinghao Zhou and Chen Wei and Huiyu Wang and Wei Shen and Cihang Xie and Alan L. Yuille and Tao Kong},
  booktitle = {International Conference on Learning Representations (ICLR)},
  year = {2022},
  litlas_id = {PW7JRR6HTEJZ},
  litlas_boards = {Foundation Model; PhiJEPA},
}

@article{oquab2024dinov2,
  title = {{DINOv2}: Learning Robust Visual Features without Supervision},
  author = {Maxime Oquab and Timothée Darcet and Théo Moutakanni and Huy V. Vo and Marc Szafraniec and Vasil Khalidov and Pierre Fernandez and Daniel Haziza and Francisco Massa and Alaaeldin El-Nouby and Mahmoud Assran and Nicolas Ballas and Wojciech Galuba and Russell Howes and Po-Yao Huang and Shang-Wen Li and Ishan Misra and Michael G. Rabbat and Vasu Sharma and Gabriel Synnaeve and Hu Xu and Hervé Jégou and Julien Mairal and Patrick Labatut and Armand Joulin and Piotr Bojanowski},
  journal = {Transactions on Machine Learning Research},
  year = {2024},
  litlas_id = {PZW3XBRQKX1R},
  litlas_boards = {Foundation Model; Computer Vision; PhiJEPA},
}

@inproceedings{he2022masked,
  title = {Masked Autoencoders Are Scalable Vision Learners},
  author = {Kaiming He and Xinlei Chen and Saining Xie and Yanghao Li and Piotr Dollár and Ross B. Girshick},
  booktitle = {IEEE/CVF Conference on Computer Vision and Pattern Recognition (CVPR)},
  year = {2022},
  litlas_id = {P2G7X66RJYN4},
  litlas_boards = {Foundation Model; PhiJEPA},
}

@article{chen2023context,
  title = {Context Autoencoder for {Self-Supervised} Representation Learning},
  author = {Xiaokang Chen and Mingyu Ding and Xiaodi Wang and Ying Xin and Shentong Mo and Yunhao Wang and Shumin Han and Ping Luo and Gang Zeng and Jingdong Wang},
  journal = {International Journal of Computer Vision},
  volume = {132},
  number = {1},
  pages = {208--223},
  year = {2024},
  litlas_id = {PX3N2GP2P46Y},
  litlas_boards = {PhiJEPA},
}

@article{balestriero2025lejepa,
  title = {{LeJEPA}: Provable and Scalable {Self-Supervised} Learning Without the Heuristics},
  author = {Randall Balestriero and Yann LeCun},
  journal = {arXiv preprint arXiv:2511.08544},
  year = {2025},
  litlas_id = {P82ZQP4G4YBQ},
  litlas_boards = {JEPA; PhiJEPA},
}

@inproceedings{dosovitskiy2021image,
  title = {An Image is Worth 16x16 Words: Transformers for Image Recognition at Scale},
  author = {Alexey Dosovitskiy and Lucas Beyer and Alexander Kolesnikov and Dirk Weissenborn and Xiaohua Zhai and Thomas Unterthiner and Mostafa Dehghani and Matthias Minderer and Georg Heigold and Sylvain Gelly and Jakob Uszkoreit and Neil Houlsby},
  booktitle = {International Conference on Learning Representations (ICLR)},
  year = {2021},
  litlas_id = {PJS2NS5403TM},
  litlas_boards = {PhiJEPA},
}

@inproceedings{ishikawa2025phinets,
  title = {{PhiNets}: Brain-inspired Non-contrastive Learning Based on Temporal Prediction Hypothesis},
  author = {Satoki Ishikawa and Makoto Yamada and Han Bao and Yuki Takezawa},
  booktitle = {International Conference on Learning Representations (ICLR)},
  year = {2025},
  litlas_id = {P800H0BAP9QR},
  litlas_boards = {Foundation Model; JEPA; PhiJEPA},
}

@techreport{krizhevsky2024learning,
  title = {Learning Multiple Layers of Features from Tiny Images},
  author = {Alex Krizhevsky},
  institution = {University of Toronto},
  year = {2009},
  litlas_id = {P02PXG9VPQWG},
  litlas_boards = {PhiJEPA},
}

@article{kumaran2016what,
  title = {What Learning Systems do Intelligent Agents Need? Complementary Learning Systems Theory Updated},
  author = {Dharshan Kumaran and Demis Hassabis and James L. McClelland},
  journal = {Trends in Cognitive Sciences},
  volume = {20},
  number = {7},
  pages = {512--534},
  year = {2016},
  litlas_id = {PG9TA5WVEY96},
  litlas_boards = {PhiJEPA},
}

@article{mcclelland1995why,
  title = {Why there are complementary learning systems in the hippocampus and neocortex: Insights from the successes and failures of connectionist models of learning and memory},
  author = {James L. McClelland and Bruce L. McNaughton and Randall C. O'Reilly},
  journal = {Psychological Review},
  volume = {102},
  number = {3},
  pages = {419--457},
  year = {1995},
  litlas_id = {P88ZX23DFR15},
  litlas_boards = {PhiJEPA},
}

@article{chen2024predictive,
  title = {Predictive sequence learning in the hippocampal formation},
  author = {Yusi Chen and Huanqiu Zhang and Mia Cameron and Terrence J. Sejnowski},
  journal = {Neuron},
  volume = {112},
  number = {15},
  pages = {2645--2658.e4},
  year = {2024},
  litlas_id = {PQT10JPM7PGX},
  litlas_boards = {PhiJEPA},
}

@inproceedings{deng2009imagenet,
  title = {{ImageNet}: A large-scale hierarchical image database},
  author = {Jia Deng and Wei Dong and Richard Socher and Li-Jia Li and Kai Li and Li Fei-Fei},
  booktitle = {IEEE Conference on Computer Vision and Pattern Recognition (CVPR)},
  year = {2009},
  litlas_id = {PQ1Y5QS438W4},
  litlas_boards = {PhiJEPA},
}

@article{yamada2025phinet,
  title = {Brain-Inspired Stochastic Joint Embedding Representation Learning},
  author = {Makoto Yamada and Kian Ming A. Chai and Ayoub Rhim and Satoki Ishikawa and Mohammad Sabokrou and Yao-Hung Hubert Tsai},
  journal = {arXiv preprint arXiv:2505.11129},
  year = {2025},
  litlas_id = {PYGREF2NRVC2},
  litlas_boards = {PhiJEPA},
}

@inproceedings{caron2021emerging,
  title = {Emerging Properties in {Self-Supervised} Vision Transformers},
  author = {Mathilde Caron and Hugo Touvron and Ishan Misra and Hervé Jégou and Julien Mairal and Piotr Bojanowski and Armand Joulin},
  booktitle = {IEEE/CVF International Conference on Computer Vision (ICCV)},
  year = {2021},
  litlas_id = {P2EDH97PEEVJ},
  litlas_boards = {PhiJEPA},
}

@inproceedings{chen2021exploring,
  title = {Exploring Simple Siamese Representation Learning},
  author = {Xinlei Chen and Kaiming He},
  booktitle = {IEEE/CVF Conference on Computer Vision and Pattern Recognition (CVPR)},
  year = {2021},
  litlas_id = {P2N0T9FVP06F},
  litlas_boards = {PhiJEPA},
}

@inproceedings{chen2020simple,
  title = {A Simple Framework for Contrastive Learning of Visual Representations},
  author = {Ting Chen and Simon Kornblith and Mohammad Norouzi and Geoffrey E. Hinton},
  booktitle = {International Conference on Machine Learning (ICML)},
  year = {2020},
  litlas_id = {PDGYHX1WP1JN},
  litlas_boards = {PhiJEPA},
}

@article{simeoni2026dinov3,
  title = {{DINOv3}},
  author = {Oriane Siméoni and Huy V. Vo and Maximilian Seitzer and Federico Baldassarre and Maxime Oquab and Cijo Jose and Vasil Khalidov and Marc Szafraniec and Seungeun Yi and Michaël Ramamonjisoa and Francisco Massa and Daniel Haziza and Luca Wehrstedt and Jianyuan Wang and Timothée Darcet and Théo Moutakanni and Leonel Sentana and Claire Roberts and Andrea Vedaldi and Jamie Tolan and John Brandt and Camille Couprie and Julien Mairal and Hervé Jégou and Patrick Labatut and Piotr Bojanowski},
  journal = {Transactions on Machine Learning Research},
  year = {2026},
  litlas_id = {PX7TX1WAP07N},
  litlas_boards = {PhiJEPA},
}

@article{bardes2024revisiting,
  title = {Revisiting Feature Prediction for Learning Visual Representations from Video},
  author = {Adrien Bardes and Quentin Garrido and Jean Ponce and Xinlei Chen and Michael G. Rabbat and Yann LeCun and Mahmoud Assran and Nicolas Ballas},
  journal = {Transactions on Machine Learning Research},
  year = {2024},
  litlas_id = {P1PZAY0TYBDM},
  litlas_boards = {JEPA; PhiJEPA},
}

@article{assran2025v,
  title = {{V-JEPA} 2: {Self-Supervised} Video Models Enable Understanding, Prediction and Planning},
  author = {Mido Assran and Adrien Bardes and David Fan and Quentin Garrido and Russell Howes and Mojtaba Komeili and Matthew Muckley and Ammar Rizvi and Claire Roberts and Koustuv Sinha and Artem Zholus and Sergio Arnaud and Abha Gejji and Ada Martin and Francois Robert Hogan and Daniel Dugas and Piotr Bojanowski and Vasil Khalidov and Patrick Labatut and Francisco Massa and Marc Szafraniec and Kapil Krishnakumar and Yong Li and Xiaodong Ma and Sarath Chandar and Franziska Meier and Yann LeCun and Michael Rabbat and Nicolas Ballas},
  journal = {arXiv preprint arXiv:2506.09985},
  year = {2025},
  litlas_id = {PXB63B76JEJ9},
  litlas_boards = {JEPA; PhiJEPA},
}

@article{yamada2026siamjepa,
  title = {{SiamJEPA}: On the Role of Siamese Student Encoders in {JEPA}},
  author = {Makoto Yamada},
  journal = {arXiv preprint arXiv:2607.04044},
  year = {2026},
  litlas_id = {P320X65AD5AZ},
  litlas_boards = {JEPA; PhiJEPA},
}

@misc{path,
  title = {A Path Towards Autonomous Machine Intelligence},
  author = {Yann LeCun},
  howpublished = {OpenReview},
  year = {2022},
  litlas_id = {browser\_import:https-scholar.google.com-citations-view\_op-view\_citation-hl-378eb6871e4ff90d},
  litlas_boards = {JEPA; PhiJEPA},
}

@inproceedings{assran2023self,
  title = {{Self-Supervised} Learning from Images with a {Joint-Embedding} Predictive Architecture},
  author = {Mahmoud Assran and Quentin Duval and Ishan Misra and Piotr Bojanowski and Pascal Vincent and Michael G. Rabbat and Yann LeCun and Nicolas Ballas},
  booktitle = {IEEE/CVF Conference on Computer Vision and Pattern Recognition (CVPR)},
  year = {2023},
  litlas_id = {P0B7XY3H57S0},
  litlas_boards = {JEPA; PhiJEPA},
}

@article{maes2026leworldmodel,
  title = {{LeWorldModel}: Stable {End-to-End} {Joint-Embedding} Predictive Architecture from Pixels},
  author = {Lucas Maes and Quentin Le Lidec and Damien Scieur and Yann LeCun and Randall Balestriero},
  journal = {arXiv preprint arXiv:2603.19312},
  year = {2026},
  litlas_id = {PG2TKAG3G4JY},
  litlas_boards = {JEPA; World model; PhiJEPA},
}

@inproceedings{he2026dseq,
  title = {{DSeq-JEPA}: Discriminative Sequential {Joint-Embedding} Predictive Architecture},
  author = {Xiangteng He and Shunsuke Sakai and Shivam Chandhok and Sara Beery and Kun Yuan and Nicolas Padoy and Tatsuhito Hasegawa and Leonid Sigal},
  booktitle = {European Conference on Computer Vision (ECCV)},
  year = {2026},
  litlas_id = {PNDFX8HJVGQQ},
  litlas_boards = {JEPA; PhiJEPA},
}

@article{dekraker2026understanding,
  title = {Understanding the hippocampus as an apex of the cortical hierarchy and self-supervised predictive learning engine},
  author = {DeKraker, Jordan and Eichert, Nicole and Hong, Seok-Jun and Evans, Alan C. and Bernhardt, Boris C.},
  journal = {Neuroscience and Biobehavioral Reviews},
  volume = {189},
  pages = {106890},
  year = {2026},
  doi = {10.1016/j.neubiorev.2026.106890}
}
